%% file: paper.tex
\documentclass{article}
\usepackage{arxiv}

\usepackage{lineno,hyperref}
\modulolinenumbers[5]

\usepackage{graphicx,natbib,amssymb}
\usepackage{algorithm}
\usepackage[noend]{algpseudocode}
\usepackage{tabularx}
\usepackage{hhline}
\usepackage{textcomp}
\usepackage{subfigure}
\usepackage{latexsym}
\usepackage{amsmath}
\usepackage{multirow}
\usepackage{multicol}
\usepackage{url}
\usepackage{booktabs}
\usepackage{subfiles}
\usepackage{todonotes}
\usepackage{enumitem}% improve itemize, enumerate, description lists
\usepackage{csquotes} %make quotation
\usepackage{textcomp} %to use aphotrophe

\usepackage{tabularx} % allow creation of custom column 

\usepackage{header}

\newcolumntype{L}[1]{>{\raggedright\arraybackslash}p{#1}} %flush left fixed width:
\newcolumntype{C}[1]{>{\centering\arraybackslash}p{#1}}  %center fixed width:
\newcolumntype{R}[1]{>{\raggedleft\arraybackslash}p{#1}} %flush right fixed width:

\newcommand{\bo}[1]{\textbf{#1}}        
\newcommand{\un}[1]{\underline{#1}}

\title{Enhanced word embeddings using multi-semantic representation through lexical chains}

\author{
  Terry Ruas\thanks{The final, published version of this article is available online in the Information Sciences Journal. Please check the final publication record for the latest revisions to this article. \url{https://doi.org/10.1016/j.ins.2020.04.048}} \\
  University of Michigan - Dearborn\\
  University of Wuppertal \\
  \texttt{ruas@uni-wuppertal.de} \\
  \And
Charles Henrique Porto Ferreira \\
Federal University of ABC\\
  \texttt{charles.ferreira@ufabc.edu.br} \\
  
   \And
William Grosky \\
  University of Michigan - Dearborn\\
  \texttt{wgrosky@umich.edu} \\
  
  \AND
Fabr\'{i}cio Olivetti de Fran\c{c}a, D\'{e}bora Maria Rossi de Medeiros \\
Federal University of ABC\\
  \texttt{\{folivetti@ufabc.edu.br,debora.medeiros@ufabc.edu.br\}} \\
  
  }

\begin{document}
\maketitle

\thispagestyle{firststyle}

% \title{Document Classification Using Multi-Semantic Embeddings Through Lexical Chains}


% \author[1,3]{Terry Ruas\corref{cor2}\corref{cor3}}
% \ead{truas@umich.edu, ruas@uni-wuppertal.de}

% \author[2]{Charles Henrique Porto Ferreira}
% \ead{charles.ferreira@ufabc.edu.br}

% \author[1]{William Grosky}
% \ead{wgrosky@umich.edu}

% \author[2]{Fabr\'{i}cio Olivetti de Fran\c{c}a}
% \ead{folivetti@ufabc.edu.br}

% \author[2]{D\'{e}bora Maria Rossi de Medeiros}
% \ead{debora.medeiros@ufabc.edu.br}

% \address[1]{University of Michigan - Dearborn, 4901 Evergreen Rd, Dearborn, MI 48128, USA}
% \address[2]{Federal University of ABC, Av. dos Estados, 5001, Santo Andr\'{e}', SP, 09210-580, Brazil}
% \address[3]{University of Wuppertal, Rainer-Gruenter-Str. 5, 42119, Germany}

%\author[mymainaddress]{\corref{mycorrespondingauthor}}

%\author[mysecondaryaddress]{National Institute of Informatics}
% \cortext[cor2]{The final, published version of this article is available online in the Information Sciences Journal. Please check the final publication record for the latest revisions to this article. \url{https://doi.org/10.1016/j.ins.2020.04.048}}
% \cortext[cor3]{\textcopyright 2020. This manuscript version is made available under the CC-BY-NC-ND 4.0 license http://creativecommons.org/licenses/by-nc-nd/4.0/}

\begin{abstract}% Shorten this 
The relationship between words in a sentence often tells us more about the underlying semantic content of a document than its actual words, individually. In this work, we propose two novel algorithms, called \textit{Flexible Lexical Chain II} and \textit{Fixed Lexical Chain II}. These algorithms combine the semantic relations derived from lexical chains, prior knowledge from lexical databases, and the robustness of the distributional hypothesis in word embeddings as building blocks forming a single system. In short, our approach has three main contributions: (i) a set of techniques that fully integrate word embeddings and lexical chains; (ii) a more robust semantic representation that considers the latent relation between words in a document; and (iii) lightweight word embeddings models that can be extended to any natural language task. We intend to assess the knowledge of pre-trained models to evaluate their robustness in document classification task. The proposed techniques are tested against seven word embeddings algorithms using five different machine learning classifiers over six scenarios in the document classification task. Our results show the integration between lexical chains and word embeddings representations sustain state-of-the-art results, even against more complex systems.
\end{abstract}

% \begin{keyword}
% Lexical chains \sep natural language processing \sep word embeddings \sep document classification\sep synsets
% %\MSC[2010] 00-01\sep  99-00
% \end{keyword}

%\linenumbers

%Introduction
\section{Introduction}\label{sec:intro}
\input{introduction.tex}

%Related Work
\section{Related work}\label{sec:relwor}
\input{relatedwork.tex}

%Core - FLLC and FXLC
\section{Semantic embeddings through lexical chains} \label{sec:lexicalchains}
\input{core.tex}

% Experiments and discussion
\section{Experimental details}\label{sec:expe}
\input{experiments.tex}

% Further Discussions and Limitations
\section{Further discussions and limitations}\label{sec:limits}
\input{limitations.tex}

%Final considerations
\section{Final considerations}\label{sec:concl}
\input{conclusions.tex}

\section{Acknowledgments}
\input{acknowledgment.tex}

%Bibliography
\section{References}
\bibliography{paper}

\end{document}

%% file: introduction.tex
%***********************************************************************************
%C: We need to link text classification content with lexical chain content. Specially the word embedding approach
%*********************************************************************************

% importance of text classification and its applicability
%                  |---> document representation 
%                             |--> bag of words and word embedding
%                                         |--> individual limitation of this two approaches 
%                                         |--> global limitation as lack of disambiguation 
%                                                         |--> lexical chain as a possible solution

%C: text classification %Text Mining~\citep{Berry:10} \citep{Drucker:99} \citep{Houvardas:06}
%C: introduce text mining and document classification
Text Mining is a sub-area of Data Mining focused on extracting knowledge from text documents automatically. Its usefulness is present in several areas in the Natural Language Processing (NLP) domain, such as detecting spam email~\citep{Heydari:15}, authorship identification~\citep{Zheng:06}, and text summarization~\citep{Liu:17a}. Among these areas, text classification has received considerable attention and has been the subject of several recent research topics~\citep{Zhang:15,Joulin:16,Ferreira:18a,Sinoara:19}. In the document classification task, the goal is to create a model using features extracted from a set of text data, often referred to as a training set, capable of inferring the correct labels of unseen text documents. The success of this model depends on how it performs on a specific task, and it generalizes to other scenarios. Feature extraction from text data is a challenge on its own in document classification.%, due to its unstructured nature.
%TR:V

%C: word embedding and its problems
%C: transfer knowledge to solve problems with the amount of data necessary - TRV
In the last few years, word embeddings-based approaches have become popular to represent words in low-dimensional spaces~\citep{Bengio:03,Mikolov_a:13,Penni:14}. The core element of these techniques relies on using neural network models to represent words according to their context (neighboring words) in a dense vector space, with each of its values corresponding to an intrinsic property of the word. Word embeddings techniques have alleviated some problems (e.g., scalability, lack of semantic relationships) of traditional count-based methods, such as bag-of-words (BOW). BOW-inspired approaches often use all the words in a corpus as features to represent documents, usually considering term-frequency and inverse-document-frequency (tf-idf) as a normalization factor in their weighting scheme. Nevertheless, neural network-based models require a considerable amount of data to accurately produce a good representation for words and their contexts, which can be troublesome for specific domains. Thus, methods that can derive results from more than one domain are beneficial.
%TR:V

%transfer learning %alternative ~\citep{Pratt:93} %REDUCE - TRV
Transfer learning presents itself as an exciting alternative to mitigate the data scarcity issue, relaxing the underlying assumption used in machine learning that requires training and test data to be from the same domain, vector space, and similar distribution~\citep{Pan:2010}. The same way humans take advantage of their previous experiences to solve future ones, transfer learning allows us to use what was learned from one task to do another. All the proposed and compared systems in this work make use of transfer learning by producing word vector representations moving from one domain to another.
%TR: V

%C: talk about lack of semantic on current data representation
One of the main challenges in representing documents is the ability to incorporate semantic aspects and to extract the relationships between words. While BOW techniques fall short in obtaining semantic knowledge from a domain, word embeddings often ignore the relationship effects of word order. One direction that tries to mitigate such issues is the construction of lexical chains~\citep{Gonzales:17,Ruas_b:17,Ruas:18}. \emph{Lexical chains} is a sequence of related words delineating portions of text for a specific topic in a cohesive manner~\citep{Morris:91}. A document often produces continuity in its ideas if its sentences are semantically connected, providing good cohesion among them. \emph{Lexical cohesion} usually occurs in words close to each other in a text, especially those adjacent to one another. %cohesive manner~\citep{Halliday:76} %lexical chains \cite{Wei:15} TRV

%our main contribution and objective with this paper - TRV
In this work, we propose two novel algorithms that combine the benefits of word embeddings, transfer learning, and lexical chains to solve the document classification. For this, we provide a model that allows us to represent any text document using the semantic relation between its words. The main idea behind lexical chains is to capture, in a more concise way, essential features to connect the meanings between words in the text. This semantic representation is obtained by one of the two proposed lexical chain algorithms, \emph{Flexible Lexical Chain II} (FLLC II) and \emph{Fixed Lexical Chain II} (FXLC II). While the former takes advantage of an external lexical database to derive its word relations, the latter defines them within a pre-defined semantic space. Besides, both proposed techniques incorporate state-of-the-art word embeddings algorithms to represent their chains. We tested our approaches on six multi-class datasets and compared our techniques against seven state-of-the-art pre-trained models over five machine learning classifiers. Our experiments show that our lexical chain semantic representation improves the overall results in five out of the six datasets. The main contributions of our work are threefold:

\begin{enumerate}
    \item Two new algorithms to extract the semantic relations between words of text documents combining the benefits of lexical chains, word embeddings, and the prior knowledge of external lexical databases.
    \item Detailed experimental validation of our techniques in the document classification task against seven different systems and six multi-class datasets.% We also provide an extensive description of all machine learning classifiers used to evaluate our models, including their best hyperparameter configurations and the processes to find them.
    \item A collection of lightweight word embeddings models (75\% smaller) and the source code for the proposed techniques are publicly available\footnote{\label{lexicalchain-github}\url{https://github.com/truas/LexicalChain_Builder}}. %so they can be used in various NLP tasks. %In addition, the source code for our algorithms is also included. 
\end{enumerate}

%paper structure TRB
%The remainder of this paper is organized as follows: Section~\ref{sec:relwor} introduces related work in word embeddings, lexical chains, and document classification. In Section~\ref{sec:lexicalchains}, our proposed techniques to build lexical chains are described in detail, including an end-to-end example. Section~\ref{sec:expe} describes the experimental details, including their results and discussions. In Section~\ref{sec:limits} we try to provide a deeper explanation of the strengths and weaknesses of our approaches while giving some alternatives that could mitigate the latter. Lastly, Section~\ref{sec:concl} presents some final considerations and future directions for this work.

%% file: relatedwork.tex
%THIS SECTION CAN CERTAINLY BE REDUCED ALONG WITH THE BIBLIOGRAPHY
%initial thoughts TRV
The idea to extract semantic information from a text and from this to construct features to refine document representations has been receiving much attention in the NLP arena. Techniques that can use the words in a document and their relations provide a more robust representation than count-based ones since they penetrate beyond the syntax layer. Moreover, systems that can combine prior knowledge, obtained from different tasks or problems, can leverage even more the representation of words and documents.

%lexical chains %Barzilay:97 text summarization TRV
The adoption of lexical chains is present in several NLP tasks, such as: word similarity~\citep{Pedersen:04}, keyword extraction~\citep{Ruas_b:17}, and document clustering~\citep{Guo:11}. More recently, some publications use lexical chains in a more specific context.~\citet{Bar:15} explore the use of text similarity metrics in the context of semantic text similarity in text re-usability detection. They argue text similarity cannot be considered a unified entity. Instead, one should evaluate at which level two documents are semantically similar or not.%~\citet{Wei:15} use lexical chains in conjunction with WordNet~\cite{Fellbaum:98,Miller:95} to extract a set of semantically connected words for the document classification task. 

%\BOW and TF-IDF + LDA  TRV
Bag-of-words (BOW) uses words as features to represent documents in a vector space model~\citep{Harris:54}. Typically, the elements' weights can be adjusted according to their relevance, such as the number of occurrences and normalized counting. Albeit BOW proposes a simple but robust approach, it has some drawbacks, such as lack of semantic information, sparsity, and scaling problems. In this context, Latent Dirichlet Allocation (LDA)~\citep{Blei:03} comes as an alternative to represent documents as a collection of topics according to their content. LDA provides a statistical model distribution over the many hidden topics in a text document based on the distribution of the co-occurrences of its words. The distributions of these words will compose the topics of the actual corpus, which is expected to represent the entire text collection in a much smaller vector space.

The robustness of context-predictive models against classic count-vector-based distributional ones has been tested in several NLP tasks and proven to be quite effective~\citep{Baroni:14}. Among the neural network models (context-predictive), some should receive special attention, such as word2vec~\citep{Mikolov_a:13}. Word2vec embeds words in a dense vector space of low dimensionality through one of the two following techniques: continuous skip-gram and continuous bag-of-words (CBOW). In the skip-gram model, one tries to predict the neighboring context of a word given a target word, while CBOW does the converse, predicting a word given its context. As influential as word2vec, GloVe~\citep{Penni:14} builds a co-occurrence matrix of the words in a corpus, to calculate the ratio between the probabilities of these words. While word2vec is focused on fixed context windows to derive its vectors, GloVe takes advantage of a global perspective. As for their results, both word2Vec and GloVe present similar findings in various NLP tasks~\citep{Iacobacci:16,Conneau:17}. %Shi:14

% lexical chains + word embeddings - TRV
Even though the applicability of lexical chains is diverse, we see little work exploring them with recent advances in NLP, more specifically with word embeddings. In~\citep{Simov:17}, lexical chains are built using specific patterns found on WordNet~\cite{Fellbaum:98} and used for learning word embeddings. Their resulting vectors, as ours, are tested in the document similarity task.~\citet{Gonzales:17} use word-sense embeddings to produce lexical chains that are integrated with a neural machine translation model. The combination of word-sense vectors and lexical chains improves their translation accuracy by 0.43\% and 0.6\% for German $\rightarrow$ English and German $\rightarrow$ French, respectively. Even though the papers mentioned use lexical chains, they do not explore them in the document classification task. We provide two novel algorithms that are built in a bottom-up approach, inspired by the \citet{Morris:91} definition of lexical chains (Section~\ref{ssec:document-to-chain}). Additionally, we derive a multi-sense embedding representation for our lexical chains and compare them with state-of-the-art word embeddings techniques. %In the same task of translation,~\citet{Mascarell:17} propose a model that uses lexical chains to leverage statistical machine translation by using a document encoder. Instead of using an external lexical database, they use word embeddings to detect the lexical chains in the source text. 

%document embeddings TRV
Extending word2vec's techniques (skip-gram and CBOW), \emph{Paragraph Vectors} (PV) is an unsupervised framework that learns continuous distributed vector representations for any size of text portion (e.g., paragraphs, documents)~\citep{Le:14}. This technique alleviates the inability of word2vec to embed documents as a unique object. Differently than word2vec, PV produces a fixed-length $n$-dimensional vector representation for each entire-textual segment, instead of just the words in the corpus.~\citet{Le:14} algorithm is also available in two different forms: Distributed Bag-of-Words of Paragraph Vectors (PV-DBOW) and Distributed Memory Model of Paragraph Vectors (PV-DM), which are analogous to the skip-gram and CBOW models in word2vec respectively. %The PV-DBOW works with the same principle as the skip-gram model, in which one tries to predict the context given a target word. In this approach, the context words are not considered as input, but instead, the model predicts words randomly sampled from the paragraph in the output. For this reason, there is no need to keep track of the word vectors of the context, causing this model to store fewer data. The PV-DM training model is similar to the CBOW approach in which the context is used to predict a target word. However, in both training models, an extra feature vector representing the text segment (called paragraph-id) is added to the context sliding window. The paragraph-id and word vectors are then averaged or concatenated to predict the target word. This forces the model to remember the missing words from the context when performing the prediction of these words. The paragraph-id vector is shared throughout the entire document for each sliding window, representing the meaning of the document as a whole. 

%other players in word embeddings more popular than doc2vec - TRV
Even though PV~\citep{Le:14} has reported superior results when compared to the original word2vec~\cite{Mikolov_b:13} approach, its impact was smaller than its predecessor in the NLP community. The widespread use of word2vec is due to, among other things, its efficient log-linear neural network language model, low-dimensional vector representation, resource-friendly implementation, and effectiveness in several NLP downstream tasks ~\citep{Iacobacci:16,Mancini:17,Ruas:19, Taher:16}. For now, we decided to compare the proposed techniques in this paper with more accessible, robust and disruptive approaches that work at least in a word-level. %Rothe:15,Li:15,

%character embeddgins TRV
Following the opposite direction of paragraph vectors, some publications moved from a direct document representation to a sub-word one. In fastText~\citep{Bojanowski:17}, they extend the skip-gram model by proposing a word representation obtained from a sum of the $n$-grams of its constituent sub-word vectors. For example, using $n = 3$, the word \emph{kiwi} would be represented as $\{<ki, kiw, iwi, wi>\}$ and the word \emph{kiwi} itself as a particular case. Adopting a similar method,~\citet{Matthew:18} propose to represent words through its constituent characters with ELMo (Embeddings from Language Models). ELMo uses a two-layer deep bi-directional Language Model (biLM) with character convolutions as a linear function of their internal states. Because of its architecture, ELMo does not keep a word-dictionary in their model, only characters. Therefore it can handle any out-of-vocabulary (OOV) words by averaging all biLM layers for the constituent characters of a word. More recently, Universal Sentence Encoder (USE)~\citep{Cer:18} proposes two encoding models to represent any text as vectors, one focused on accuracy  (Transformer architecture) and the other on inference (Deep Averaging Network). The former builds a sub-graph that uses attention mechanisms to compute the context representations of words in a text, considering their ordering and identity. The latter average words and bi-grams embeddings and feed them in a feed-forward deep neural network to produce new embeddings. In Section~\ref{sec:expe}, we compare our proposed techniques, trained in a simple word2vec implementation, against the aforementioned state-of-the-art models.

One limitation with current word embeddings techniques is they condense all meanings of a word in single vector representations. Traditional algorithms rely on the context of the words to help express their multiple purposes in the $n$-dimensions of their vectors. An alternative to mitigate such a problem is to use labeled word-senses in a corpus to train word embedding models. However, labeled word-senses on a large scale are scarce, and human disambiguation is expensive, time-consuming, and subjective. Recently,~\citet{Ruas:19} proposed a new technique called \emph{Most Suitable Sense Annotation} (MSSA) that performs unsupervised word-sense disambiguation and annotation in words from any part-of-speech (POS), which are later used in a word embeddings algorithm to improve the quality of traditional word embeddings model generation.~\citet{Ruas:19}'s technique finds the most suitable word-sense using the word's context, the semantic relationships expressed in WordNet~\cite{Fellbaum:98}, and a pre-trained word embeddings model (e.g., Google Vectors~\cite{Mikolov_a:13}). In addition to the vanilla MSSA technique, the authors also presented two other variants of their approach, named MSSA-1R and MSSA-2R. Their objective is to refine the word embeddings model from MSSA through an iterative process. The MSSA algorithm and its variations were tested in a word similarity task in six different benchmarks against a large number of systems, producing state-of-the-art results~\citep{Ruas:19}. The lexical chain algorithms proposed in this paper (Sections~\ref{ssec:fllc} and~\ref{ssec:fxlc}) are built using the disambiguation process from MSSA and their multi-sense word embeddings models, which are all publicly available\footnote{\url{https://github.com/truas/MSSA}}.
 
%text classification using embeddings TRV
In \citep{Fu:18}, they propose a technique to generate document representations called Bag-of-Meta-Words (BoMW).  In this technique, they use pre-trained word embeddings models (e.g., word2vec, GloVe) to retrieve and average the vector representation of the constituent words of a document. They map each vector into a different feature space and sum the vectors from this new space to represent a document. They propose two approaches, namely the Naive Interval Meta-Word Model and Features Combination Meta-Word Model. The former discretizes features from the initial word embedding, while the latter clusters related features under the same group. They compare their results on three datasets against BOW, neural networks models (e.g., recurrent neural networks), and an average of word embedding models. Their approach shows an improvement in accuracy over the baselines on the document classification task.

Some techniques make use of context information in the document, to incorporate semantic aspects into traditional word embeddings models. In~\citep{Kamkarhaghighi:17}, they build a content tree word embedding to capture local information from the words in a document. Their word vectors are calculated as a weighted average of a word's vector and its immediate parent. They assume that the context, represented by a word parent node, influences its neighboring words. The insertion of a new word into an existing content tree only happens if there is a high correlation between all nodes in that tree and the new word.~\citet{Enriquez:16} explore the complementary information of BOW and word2vec to represent their document. Their technique uses a simple voting system that averages the confidence values returned by BOW and word2vec to classify a text either in a negative or positive class. They concluded that BOW provides the best representation over word2vec, while their combination improves the overall results.

%TR closing/comparing sentence - TRV
To the best of our knowledge, this is the first work that combines lexical chains and word embeddings applied to the document classification problem. We expect the proposed algorithms to produce a robust semantic representation through the use of lexical chains. Furthermore, we build our lexical chains using several synset objects\footnote{\label{wordnet_atributte_list}\emph{hypernyms, instance\_hypernyms}, \emph{hyponyms}, \emph{instance\_hyponyms}, \emph{ member\_holonyms}, \emph{substance\_holonyms}, \emph{part\_holonyms},\emph{ member\_meronyms}, \emph{substance\_meronyms}, \emph{part\_meronyms}, \emph{attributes}, \emph{entailments}, \emph{causes}, \emph{also\_sees}, \emph{verb\_groups}, \emph{similar\_tos}, \emph{topic\_domains}, \emph{region\_domains}, and \emph{usage\_domains}} in the lexical database in addition to hypernyms and hyponyms, which are commonly used in the literature~\citep{Gonzales:17, Mascarell:17, Simov:17}  (detailed in Section \ref{sec:lexicalchains}). The main idea is to bring the semantic relations of lexical chains to traditional word embeddings techniques, leveraging their vector representation, and improving the overall result in the document classification task.

%% file: core.tex
In this work, we propose a novel approach to capture the semantic relationship between tokens from a text. Our techniques combine the use of word embeddings, lexical chains, and the prior knowledge of lexical databases, to derive the relations between words. This is done through two algorithms: \emph{Flexible Lexical Chains II (FLLC II)} (Section~\ref{ssec:fllc}) and \emph{Fixed Lexical Chains II (FXLC II)} (Section~\ref{ssec:fxlc}). Both algorithms are inspired by the approaches proposed in ~\citep{Ruas_b:17,Ruas:18}. While the lexical chains explored in~\citep{Ruas_b:17,Ruas:18} consider only one part-of-speech (POS) (nouns) and \emph{hypernyms}, ours, on the other hand, are able to deal with any POS tag, incorporate word embeddings, and also include other 19 lexical objects\textsuperscript{\ref{wordnet_atributte_list}} from the English WordNet\footnote{\label{wordnet_atributte}\url{https://www.nltk.org/_modules/nltk/corpus/reader/wordnet.html}},~\citep{Fellbaum:98}. As a result, we leverage the semantic representation of words, sentences, paragraphs, and entire documents through the use of lexical chains.  %TRV

\begin{figure}[htb]
    \centering
    \resizebox{\textwidth}{!}{
    \includegraphics[]{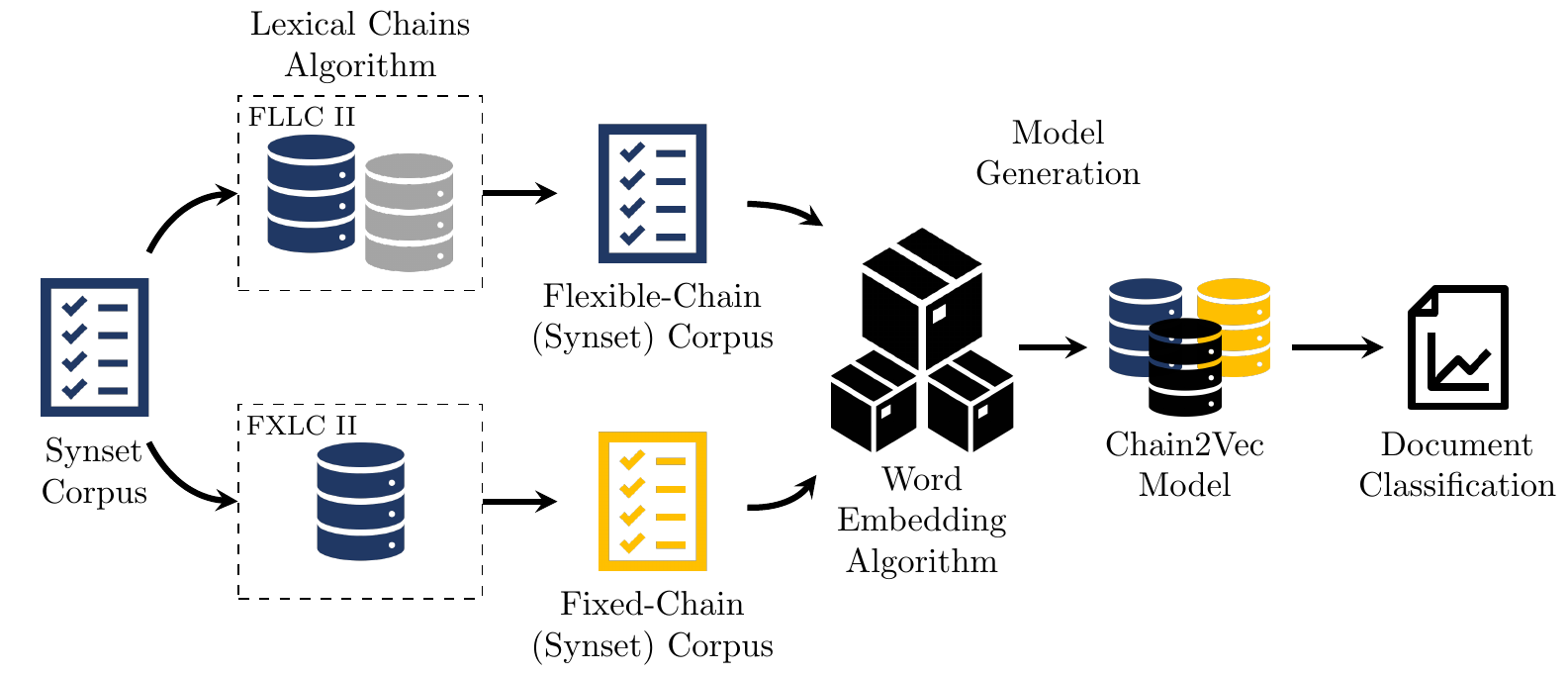}
    }
    \caption{Overview architecture for building lexical chains}
    \label{fig:lc-arch}
\end{figure}

We illustrate an overview of the entire process in Figure~\ref{fig:lc-arch}. In both cases, FLLC II and FXLC II, the constructed chains are represented by one of their constituent elements, which is selected by considering their vector representation in a pre-trained token embeddings model (e.g., word2vec). Later, we feed the output of our techniques, i.e., the synset-annotated corpus, into a word embeddings algorithm, which produces a lexical chain embeddings model. Later, the synset-vectors are used as features in the document classification task. %TRV

%[TR] - Recurrent approach used as future work:  One of the many advantages of our system is that once we produce a word embeddings model using the output of our approaches, we can use this same model to refine the representation of our chains recurrently. 

FLLC II and FXLC II techniques are, at their core, reducing the number of valid elements (words) in a document. With fewer features, our approaches can reduce the number of necessary tokens to represent a document and still keep their semantic content.  Given the characteristics mentioned above, we can classify the FLLC II algorithm as a soft-dimension reduction technique, while FXLC II as a hard-dimension reduction. Another critical difference between the proposed methods is their databases and how they are used. FLLC II chains are generated using a lexical database (e.g., WordNet) and a pre-trained token embeddings model represented by the gray and blue figures, respectively. While the former is responsible for providing the semantic relationship between the tokens in the document, the latter helps us to decide which token will represent our chain. In both algorithms, the blue figure represents a trained synset embeddings model\footnote{A word embedding algorithm using an annotated-synset corpus.}. In the FXLC II technique, since we do not group tokens directly by their semantic affinity, we do not require a lexical database with word relationships, such as WordNet~\cite{Fellbaum:98}. Instead, we need a compatible pre-trained token embedding model, concerning the document processed, to build our chains. Therefore, FXLC II can be exported to other scenarios using any pre-trained token embeddings model (e.g., words, synsets), as long as it contains the elements from the documents. In the next sub-sections, both techniques are explored in-depth, followed by a simple toy example illustrating their operation. %TRV

\subsection{From document to lexical chains} \label{ssec:document-to-chain}
%external thesaurus (Roget's Thesaurus~\citep{Rogets:79})
Lexical chains are built according to a series of relationships between words in a text document. In the seminal work of \citet{Morris:91} they consider an external thesaurus as their lexical database to extract these relations. A lexical chain is formed by a sequence of words $\{w_{1}, w_{2}, \ldots, w_n\}$ appearing in this order, such as any two consecutive words $w_i, w_{i+1}$ possess the following properties\footnote{Where category, indexes, and pointers are attributes in the lexical database considered.} \cite{Morris:91}: (a) two words share one common category in their index; (b) the category of one of these words points to the other word; (c) one of the words belongs to the other word's entry or category; (d) two words are semantically related; and (e) their categories agree to a common category.

%\begin{itemize}%CAN YOU REMOVE SOME SPACE BETWEEN THE LIST ITEMS?
%  \item Two words share one common category in their index;
%  \item The category of one of these words points to the other word;
%  \item One of the words belongs to the other word's entry or category;
%  \item Two words are semantically related;
%  \item Their categories agree to a common category.
%\end{itemize}

%TRV
%As for the quality of a built chain, three factors should be considered to measure how strong a lexical chain is \citep{Morris:91}: \textit{reiteration}, \textit{length}, and \textit{density}.\textit{ Reiteration} shows how often a lexical chain occurs, the \textit{length} is the number of tokens on it, and \textit{density} is the ratio of words in the lexical chain to the words in the text document. Our main objective is to evaluate how the proposed lexical chains perform in a downstream task, so we leave the evaluation of their strength to future work. %Thus, we consider the five characteristics defined by \citet{Morris:91} to build our lexical chains and validate their quality in the document classification problem.  %TRV

%TRV
The main goal of FLLC II and FXLC II is to represent a collection of words by their semantic values more concisely. Even though FLLC II and FXLC II outputs format are the same, they explore different aspects in capturing the lexical cohesion in a text. In FLLC II, the semantic sets (lexical chains) are assembled dynamically according to the semantic content for each token evaluated and the relationship with its adjacent neighbors. As long as there is a semantic relation that connects two or more words, they should be combined into a unique concept. If a word without any semantic affinity with the current chain presents itself, we must start a new lexical chain to capture this new idea. On the other hand, in FXLC II, text documents are broken down into pre-defined chunks, with $C$ words each, describing their semantic content. Different from FLLC II, the FXLC II technique groups a certain number of words into the same structure, regardless of their semantic affinity.

%OLD The main goal of FLLC II and FXLC II is to represent a collection of words by their semantic values in a more concise and robust way. Even though FLLC II and FXLC II outputs format are the same, they explore different aspects in capturing the lexical cohesion in a text. In the FLLC II version, the semantic sets (lexical chains) are assembled dynamically according to the semantic content for each token evaluated and the relationship with its adjacent neighbors. As long as there is any semantic affinity (i.e. a semantic relation that connects two words) they should be integrated into one single entity that represents a single concept. If a word without any semantic affinity with the current chain being built presents itself, a new lexical chain must be started so a new concept can be captured. On the other hand, in the FXLC II approach text documents are broken down into pre-defined chunks, with $C$ words each, that will hopefully describe the semantic content of them. Different from the FLLC II algorithm, the FXLC II technique groups a certain number of words into the same structure, regardless of the existence of semantic affinity or not.

%[TR] - pre-req of both FLLC and FXLC%CHANGE "..WE MAKE USE OF THE MSSA ALGORITHM" TO "..WE MAKE USE OF OUR MSSA ALGORITHM"
A requirement for both, FLLC II and FXLCII, is a disambiguated and synset-annotated corpus. We build our lexical chains according to how the word-senses (synsets) are related to each other in a sentence. For this reason, we make use of the MSSA algorithm~\cite{Ruas:19} to transform our word-based corpus to its word-sense version composed of synsets. We also compared our approach against MSSA in our experiments (Section~\ref{sec:expe}), so we can evaluate how our lexical chains compare with a state-of-the-art technique that produces multi-sense embeddings. Nonetheless, our proposed algorithms can be applied to any method that can disambiguate and annotate words according to their word senses. %TRV
%OLD An important requirement for lexical chain algorithms proposed in this paper (FLLC II  and FXLC II) is that both of them need, as input, a disambiguated and annotated synset corpus. This is because our lexical chains are built according to how the word-senses (synsets) are related to each other in a sentence. For this reason, we make use of the MSSA algorithm~\cite{Ruas:19} to transform our word-based corpus to its word-sense version composed of synsets. We also compared our approach against our MSSA algorithm in our experiments (Section~\ref{sec:expe}) so we can evaluate how our lexical chains compare with a state-of-the-art technique that produces multi-sense embeddings. Nonetheless, any technique that is able to disambiguate and annotate word according to their word sense can also be used.

%FLLC II Algortihm
\subsubsection{Flexible Lexical Chains II (FLLC II)}\label{ssec:fllc}
The FLLC II algorithm works building semantic sets of tokens that present any level of semantic relation between them. The decision of incorporating a new word into a chain is dynamic and based on 19 lexical objects\textsuperscript{\ref{wordnet_atributte_list}} in WordNet~\citep{Fellbaum:98}.% (cf. Section~\ref{sec:lexicalchains}).

We extend~\citet{Ruas_b:17} flexible chain algorithm to use all POS and also incorporate word embeddings containing the vector representation of the tokens in a set of documents $d$. In its previous version, FLLC I was designed to work in the keyword extraction problem, could not handle POS different than nouns and was not able to incorporate word embeddings. As illustrated in Algorithm~\ref{alg:fllc}, we require a set of documents $d$ represented by synsets (i.e. set of synonyms)\footnote{\url{https://wordnet.princeton.edu/documentation/wngloss7wn}}, a lexical database $ld$, and a pre-trained synset embeddings model $tsm$\footnote{\url{https://github.com/truas/MSSA}}. To transform a word-based-document into a synset-based one, we apply the MSSA algorithm proposed by~\citet{Ruas:19}. MSSA uses Google News vectors\footnote{\url{https://code.google.com/archive/p/word2vec/}} to disambiguate words with multiple senses in a text. MSSA produces the most appropriate word-sense representation of a word given its immediate context window. A context-window considers three words at a time, i.e., former $i-1$, center $i$, and latter $i+1$, in an overlapping manner, and uses the average vector of the words from the glosses to represent each word-sense. The word-sense from $i$, with the highest cosine similarity concerning its neighbors ($i-1$ and $i+1$), is selected to represent the central word $i$ in the context window. More details about MSSA and its alternatives are provided in \cite{Ruas:19}. %TRV

Once the synset embedding models are available, we can feed the system again, using the output vectors from previous passes, improving the disambiguation step in MSSA. Using a synset embedding model allows us to refine our representation and remove FLLC II and FXLC II from Google News vectors' dependency. We call this approach MSSA-NR, where $N$ is the number of feedback iterations used. None of the related work (Section~\ref{sec:relwor}) nor compared systems in our experiments (Section~\ref{sec:expe}) explore this recurrent characteristic in their systems. %TRV
% In a few words, the MSSA algorithm works by disambiguating a given word, with respect to its word-senses and considering the effect of its immediate neighbors. %TR - I will move this explanation before the algorithms to explain it bretter

Once we have a document represented as synsets, we build our flexible chains from the first to the last token. We start our current chain using $S_{1}$ (first synset in the document) to initialize the synset list used to represent the current chain $current\_chain.synsets$, and a set of related synsets that map the semantic relation between consecutive synsets $current\_chain.related$ (lines~\ref{fllc:00}:\ref{fllc:01}). The synsets retrieved in $get\_related\_syns(S_{i},ld)$ (including $S_{i}$) are a collection of extracted synsets from 19 attributes\textsuperscript{\ref{wordnet_atributte_list}} in the lexical database $ld$ (WordNet). %TRV

For each new synset evaluated, $S_{i}$, we extract their related synsets (including $S_{i}$), called $new\_rel$, and verify if there are any common synsets with the related synsets in the chain being built ($current\_chain.related$) (lines~\ref{fllc:02}:\ref{fllc:03}). In case the intersection between $current\_chain.related$ and $new\_rel$ is not empty, we add $S_{i}$ and $new\_rel$ to $current\_chain.synset$ and $current\_chain.related$ respectively (lines~\ref{fllc:04}:\ref{fllc:05}). Otherwise, it means there are no common related synsets between the current chain and $S_{i}$, so, we understand the current chain must be properly represented and added to the list of flexible chains (line~\ref{fllc:06}). Thus, we find the synset with the highest cosine similarity against the average of all synset vectors in $current\_chain.synsets$. The average of all synsets in $current\_chain.synsets$ is calculated considering a pre-trained synset embeddings model ($tsm$) for the function $get\_best\_rep(current\_chain.synset,tsm)$. The synset embedding model ($tsm$) used is produced using a Wikipedia (English) dump from 2010 \citep{Westbury:10} annotated into synsets with MSSA~\citep{Ruas:19}. After representing and including the current chain in the list of flexible chains, we start to build a new chain, but now considering $S_{i}$ (lines~\ref{fllc:07}:\ref{fllc:08}), the same method we use to start our algorithm. %TRV

After we iterate over all synsets in $d$, we also verify if there are elements in the current chain not added to our flexible chains list (line~\ref{fllc:09}). This step mitigates the problem in which all synsets are combined in one single chain or the last synset $S_{n}$ in the document is semantically related to the current chain being built. At the end of the FLLC II algorithm, we return a list of synsets representing all the lexical chains found in a document $d$ (line~\ref{fllc:10}). %TRV

\begin{algorithm}[h]
\small
   \caption{Flexible Lexical Chain II Algorithm (FLLC II)} \label{alg:fllc}
    \begin{algorithmic}[1]
    
    \Require $d = \{S_{1}, \ldots, S_{n}\} : S_{i} \in ld$; $tsm = $ trained synset embedding model; $ld = $ lexical database
    \Function{FLLC II}{$d$, $tsm$, $ld$}
        \State flexible\_chains\_list = $\emptyset$
        \State current\_chain.$synsets$ = $[S_{1}]$ \label{fllc:00}
        \State current\_chain.$related$ = $\{$get\_related\_syns($S_{1}$, $ld$)$\}$ \label{fllc:01}
        
        \For{$i$ = 2 to $n$} \label{fllc:02}
            \State new\_rel = $\{$get\_related\_syns($S_{i}$, $ld$)$\}$
            
            \If {current\_chain.$related$ $\cap$ new\_rel \textbf{not} $\emptyset$}\label{fllc:03}
                \State \textbf{Add} $S_{i}$ \textbf{to}  current\_chain.$synsets$ \label{fllc:04}
                \State \textbf{Add} new\_rel \textbf{to}  current\_chain.$related$ \label{fllc:05}
            \Else
                \State \textbf{Add} get\_best\_repr($current\_chain.synset$, $tsm$) \textbf{to} flexible\_chains\_list \label{fllc:06}
                \State current\_chain.$synsets$ = $[S_{i}]$ \label{fllc:07}
                \State current\_chain.$related$ = $\{$get\_related\_syns($S_{i}$, $ld$)$\}$ \label{fllc:08}
            \EndIf 
        \EndFor
        \If { current\_chain.$synsets$ \textbf{not} $\emptyset$} \label{fllc:09}
            \State \textbf{Add} get\_best\_repr($current\_chain.synsets$, $tsm$) \textbf{to} flexible\_chains\_list
        \EndIf
        \State \textbf{return} flexible\_chains\_list \label{fllc:10}
    \EndFunction
\end{algorithmic}
\end{algorithm}

%talk how we improved previous work
The proposed algorithm for flexible chains improves its predecessor \citep{Ruas_b:17} in several aspects. We consider all POS when building our chains, instead of just nouns. We also consider 19 attributes\textsuperscript{\ref{wordnet_atributte_list}} in WordNet for each synset evaluated, which is an improvement from most lexical chain systems that often focus on hypernyms and hyponyms only. In our version of lexical chains, we use the transfer knowledge from an external training task of word embeddings learned from the entire English Wikipedia. %TRV

\subsubsection{Fixed Lexical Chains II (FXLC II)} \label{ssec:fxlc}
As in the FLLC II algorithm (Section~\ref{ssec:fllc}), FXLC II also builds its chains using a list of synsets. However, the FXLC II algorithm has a more general approach. The lexical chains in FXLC II are defined beforehand and do not require an explicit semantic relation between its synsets. In other words, we enforce the number of synsets for each chain (chunk) throughout the document instead of building each chain according to an existing semantic relation. %TRV

We extend the~\citet{Ruas:18} algorithm for fixed chains to all POS and incorporate word embeddings containing the vector representation of the tokens in our document. In its previous version, FXLC I was tested in a small scenario with only 30 documents, it required a corpus composed exclusively of nouns, and it did not account for word embeddings in its structure. To maintain consistency between our algorithms and experiments, we also use the MSSA algorithm \citet{Ruas:19} to produce a list of synsets out of a given document $d$. Once we have a document $d$ represented as synsets, we create a new document representation called a $chunked\_document$, which is composed of chunks $C_{j}$ of size $cs$. %(line~\ref{fxlc:01}). % called $C_{j}$, where $C_{j}$ $=$ \{$S_{p}$,...,$S_{k}$\} and $S_{p}$ $\in$ $C_{j}$ (line\ref{fxlc:core:1}). As for $m$ and $k$, they are the maximum number chains throughout the text and the maximum number of synsets in each chain, respectively.
Analogous to Algorithm~\ref{alg:fllc}, we also find the synset to represent each fixed lexical chain $C_{j}$ in $chunked\_document$ through the synset with the highest cosine similarity against the average of all synset vectors in $C_{j}$. %For that, we consider a pre-trained synset embeddings model ($tsm$) for the function $get\_best\_repr(C_{j},tsm)$. We add each synset $S_{i}$ in $C_{j}$, for the highest cosine similarity value, to our list of fixed chains, which is returned by the end of FXLC II algorithm (line~\ref{fxlc:03}). %TRV

%C simplified version
%\begin{algorithm}[H]
%\small
%   \caption{Fixed Lexical Chains II Algorithm (FXLC II)} \label{alg:fxlc}
%    \begin{algorithmic}[1]
%    \Require $d = \{S_{1}, \ldots, S_{n}\} : S_{i} \in ld$; $cs =$ chunk size for each lexical chain; $tsm =$ trained synset embedding model
%    \Function{FXLC II}{$d$, $cs$, $tsm$}
%        \State chunked\_document = split($d$, $cs$) \label{fxlc:01}
%        \State fixed\_chains\_list = $\emptyset$
%        \For{ $j$ = 1 to length($chunked\_document$)} 
%            \State \textbf{Add} get\_best\_repr($C_{j}$, $tsm$) \textbf{to} fixed\_chains\_list \label{fxlc:02}
%        \EndFor
%    \State \textbf{return} fixed\_chains\_list \label{fxlc:03}
%    \EndFunction
%\end{algorithmic}
%\end{algorithm}

%how we are better than the previous work
The proposed technique for fixed chains also surpasses its predecessor \citep{Ruas:18} in some aspects. As in FLLC II, we also consider all POS when building our chains. Besides, the FXLC II technique does not rely on distance measures to calculate how far our synsets are from each other to represent each chunk~\citep{Meng:13}. Instead, based on the information provided from the synset vectors we find the closest semantic candidate in a chain for all its inner elements. %TRV  %Thus, making FXLC II dependent only on the used corpus (document) and a compatible pre-trained word embeddings model.
% \citep{Tversky1977,Lesk:86,Palmer:94, Resnik:95,Jiang:97,Leacock:98,Lin:98}

Since the FXLC II approach ignores the direct semantic affinity between synsets and groups them for each chunk $C_{j}$, this approach can be extended to other document representations as well. As long as the pair document-tokens and embedding models have the same representation, FXLC II can be applied. %We leave this investigation for future projects. %TRV

\subsection{From lexical chains to embeddings (Chains2Vec)}\label{ssec:lc2vec} %HEREREREERERER
After transforming our documents into lexical chain representations, which are formed by synsets, we use them as a training corpus input in word2vec. As a result, word2vec uses these annotated documents to produce an embedding model based on synsets. The main idea of FLLC II and FXLC II is to obtain a better semantic representation for words from a large collection of documents that generalizes well enough to any NLP downstream task or problem. 

In order to keep our vectors easy to interpret to other systems using synsets, we represent each token in our corpus using the same format as the one in \citet{Ruas:19} when applying the MSSA technique: $word\#synset\_offset\#pos$, where $word$ is the word itself, normalized in lowercase; $synset\_offset$ is an 8 digit, zero-filled decimal integer that corresponds to a unique word-sense, and $pos$ is a part-of-speech tag (e.g. $n$ for nouns, $v$ for verbs, $a$ for adjective, $s$ for adjective satellite and $r$ for adverb)\footnote{\url{https://wordnet.princeton.edu/documentation/wndb5wn}}. Since we are using an equivalent notation, the synset embeddings models produced with FLLC II and FXLC II results can be incorporated in other systems with the same format as well. %TRV

For now, we currently explore the document classification task. Still, we believe our chains can be applied to other domains the same way we used MSSA synset models \citep{Ruas:19} in our experiments. MSSA was designed for the word similarity task, but, in this work, we use and extend its synset models to the document classification task. In the following section, we provide a toy example of how FLLC II and FXLC II are used to produce our lexical chains. %TRV

%toy example for how lexical chains work

\subsection{Building lexical chains} \label{ssec:example}
\begin{figure}[!t]
    \centering
    \resizebox{\textwidth}{!}{
    \includegraphics{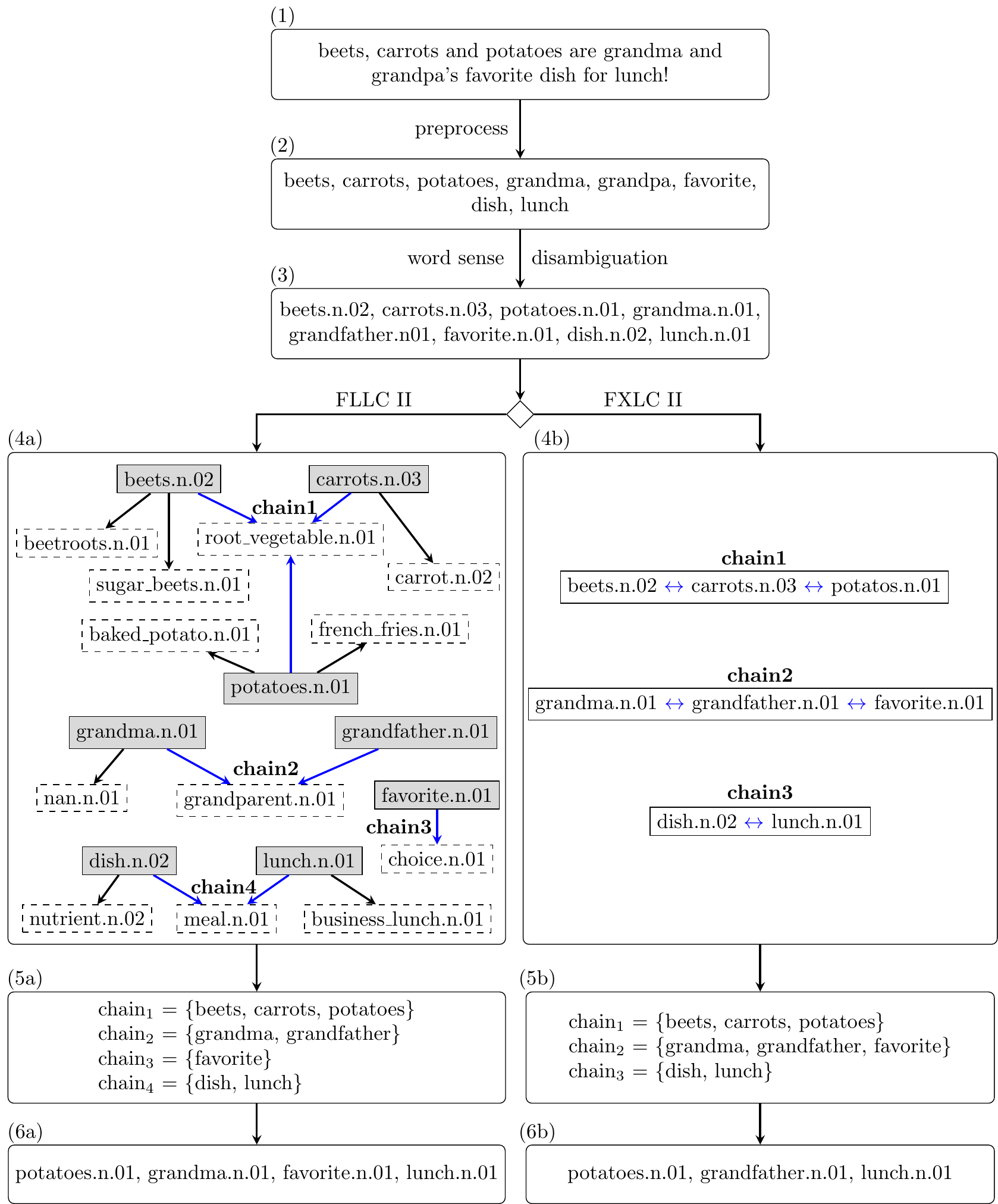}
    }
    \caption{Practical example applying FLLC II and FXLC II algorithms on a single document.}
    \label{fig:toy_example}
\end{figure}

%New toy example - intro
In this section, we provide an illustrative example for the FXLC II and FLLC II algorithms, as Figure~\ref{fig:toy_example} shows. 
%Considering the algorithms introduced in Sections~\ref{ssec:fllc} and \ref{ssec:fxlc} we provide an illustrative example of how our lexical chains are built, as Figure~\ref{fig:toy_example} shows. 

Let us consider the sentence: \textit{Beets, carrots, and potatoes are grandma and grandpa's favorite dish for lunch!}(1). As explained before, we require a document composed of synsets to build our lexical chains. After cleaning our sentence example (2), by lowercasing all the words, removing all punctuation and common English stop-words, we apply the MSSA algorithm~\citep{Ruas:19} (3), to obtain the proper synset for each word. Next, we use either the FLLC II or FXLC II.

%FLLC II
For the FLLC II, we first extract all related synset for each synset in the sentence and group the ones with at least one overlap between them ((4a) - highlighted connections). The gray square represents the document synsets, and the dashed squares represent the related synsets. In the flexible version of the algorithm, we build the chains dynamically, so as long there is a common synset between two consecutive tokens, they will be part of the same chain. If there are no overlapping synsets in the documents, the chain will have just one element, i.e., the synset itself (\textbf{chain3}). Later, from a pre-trained synset model, we extract the vectors of the synset in each chain and calculate their average (centroid) (5a). Our final document representation will have only one synset per chain. Therefore, we select the closest synset-vector to its centroid according to their cosine similarity. We hypothesize that the closer a synset is to a chain centroid, the better semantic representation it provides (6a).

%FXLC II
For FXLC II, we first define how many synsets each chunk will have and group them sequentially (4b). In this example, we are using a chunk-size of 3. In case there are fewer synsets than the actual chunk-size, then all synsets are included (\textbf{chain3}). Once the chains are formed (5b), we select one synset per chain as we do with the FLLC II algorithm. We average the vectors and select the synset closest to its centroid (6b). 

%% file: experiments.tex
%THIS SECTION SHOULD BE SQUEEZED A BIT. IS IT POSSIBLE TO LEAVE LEAVE SOME DATA SETS OUT OF TABLE 11 AND JUST POINT OUT THE RELEVANT RESULTS? SAME WITH FIGURE 2.
In this section, we explain all the details and constraints of our experiments. We first describe the different aspects of each dataset used, their characteristics, references, and availability. Next, we provide an accurate description of the chosen metrics, machine learning classifiers, and hyperparameters adopted. We include the procedures to fine-tune all experiments, classifiers, and discuss the main aspects of state-the-art systems compared to our proposed techniques. %TRV

\subsection{Datasets Details} \label{ssec:dataset}

Our experiments considered 6 different datasets with specific characteristics which impose a particular challenge on each classification. Among these datasets, 4 of them (Ohsumed\footnote{\url{http://disi.unitn.it/moschitti/corpora.htm}{http://disi.unitn.it/moschitti/corpora.htm}\label{note1}}, 20NewsGroup\footnote{\url{http://qwone.com/~jason/20Newsgroups/}{http://qwone.com/~jason/20Newsgroups/}}, Reuters\textsuperscript{\ref{note1}}and BBC\footnote{\url{http://mlg.ucd.ie/datasets/bbc.html}{http://mlg.ucd.ie/datasets/bbc.html}}) represent benchmark datasets widely used on text classification problems and the other 2 (ScyGenes and ScyClusters) are formed by scientific papers abstracts in biology. Table \ref{tab:dataset_description} describes the main characteristics of each dataset. %TRV % put the datasets names with footnotes.

%C: double check the content of this table
\input{tables/table_datasets.tex}

%C talk about general characteristics of the datasets
Table~\ref{tab:dataset_description} shows the details of each dataset concerning their theme, number of documents, classes, tokens, and synsets. We use the term \emph{token} instead of words because some features do not represent a proper word itself. As an example, we have the token \emph{housd} that does not constitute an English word, but it is present in the dataset. Typos and malformed word-tokens are not able to be processed by techniques relying on proper syntax, but recent word embeddings models such as ELMo~\citep{Matthew:18} and USE~\citep{Cer:18} can handle these issues. Column $\#synsets$ shows the number of synsets generated applying the MSSA algorithm~\citep{Ruas:19}. As a result, the number of synsets and tokens are different, and in most cases, it is smaller. Not every token (e.g., \emph{housd}) has a mapping to WordNet, which forces their exclusion since no semantic relation available. %TRV

\subsection{Machine learning classifiers}
%C classifiers used
We considered five classifiers in our experiments. These classifiers were chosen among the most popular ones in the document classification arena: \emph{K}-Nearest Neighbors (\emph{K}-NN), Support Vector Machine (SVM), Logistic Regression (LR), Random Forests (RF), and Naive Bayes (NB). %TRV

%\emph{K}-Nearest Neighbors (\emph{K}-NN)~\citep{Aha:91}, Support Vector Machine (SVM)~\citep{Cortes:95}, Logistic Regression (LR)~\citep{Mccullagh:89}, Random Forests (RF)~\citep{Breiman:01}, and Naive Bayes (NB)~\citep{Mitchell:97}.

%C talk about grid search
%C cross validation + %statistical significance
All classifiers had their parameters fine-tuned to provide the best configuration for each technique. For this reason, we performed an extensive grid search on the training corpus to find out the most suitable hyperparameters for each classifier. Except for Reuters-21578 and 20Newsgroups, all data sets do not include a training and test split, so we applied \emph{k}-cross-validation for \emph{k} equal to 10. The classification performance for all machine learning classifiers and datasets was validated in the Friedman test and Nemenyi's posthoc test for a $p$-value of less than 0.05, thus certifying the statistical significance of our experiments~\cite{Demsar:06}. %TRV
%Table~\ref{tab:gridsearch} shows the grid search configuration considered. Each classification was performed considering the best hyperparameters found.

%C evaluation metric applied
We evaluate our models in the text classification task through their classification F1-Micro score as a metric. We also generated the accuracy and F1-Macro scores for all experiments. However, their variation was minimal, almost all values of accuracy and F1-Micro were the same, and the differences between these metrics followed the same delta among different systems. Due to space constraints and the aspects above, we decided to report just the F1-Micro score.%TRV % as shown in Equation \ref{eq:accuracy}.

%\input{tables/gridsearch_configuration.tex}

% Word embeddings - models explanation
\subsection{Word embedding models characteristics} \label{ssec:wordembeddings_details}
This section presents the main characteristics of the compared systems used in the document classification task. We compared our techniques against seven state-of-the-art approaches which incorporate transfer learning at some level in their implementation. Table~\ref{tab:model_details} summarizes the algorithms used in our experiments, named: Latent Dirichlet Allocation (LDA)~\citep{Blei:03}, word2vec (w2v)~\citep{Mikolov_a:13}, Global Vectors (GloVe)~\citep{Penni:14}, fastText~\citep{Bojanowski:17}, Universal Sentence Encoder (USE)~\citep{Cer:18}, Embeddings from Language Models (ELMo)~\citep{Matthew:18}, MSSA-1R (-1R)~\citep{Ruas:19}, FLLC II using the MSSA-1R synset embeddings model (FL-1R), and FXLC II with a chunk size equal to 2 and using the MSSA-1R embeddings model (FX2-1R). %TRV %The approaches using the base MSSA embeddings models are marked with -0R next to them.  

%word embeddings systems details
\input{tables/table_modeldetails.tex}

For each system, we transformed each word in the documents into their vector representation according to an external pre-trained word embedding model. Once we obtained these word vectors, we averaged them to represent the entire document as one single entity. The systems compared in our experiments can be categorized into two major groups: (i) pre-trained word embeddings models and (ii) explicitly trained word embeddings models. In (i), we use out-of-the-shelf pre-trained models: word2vec\footnote{\url{https://code.google.com/archive/p/word2vec/}}, GloVe\footnote{\url{https://nlp.stanford.edu/projects/glove/}}, fastText\footnote{\url{https://fasttext.cc/docs/en/english-vectors.html}}, USE~\footnote{\url{https://tfhub.dev/google/universal-sentence-encoder/2}}, ELMo\footnote{\url{https://tfhub.dev/google/elmo/2}}, and MSSA (-1R and -2R)\footnote{\url{https://github.com/truas/MSSA}}. For (ii), we trained the embedding models from scratch for the following techniques: LDA, FL-1R, and FX2-1R. All techniques have a 300 dimension word vector representation, except for USE and ELMo, which are only available in 1024 and 512 dimensions, respectively. %TRV

%Put the details on the trained word embeddings and doc embeddings algorithms.
For (ii), we produced our multi-sense embeddings using the same hyperparameter configuration in their word2vec training phase: CBOW training model, 300 dimensions, a window size of 15, the minimum word count of 10, and hierarchical softmax. Any attribute not mentioned was set to its default value according to the \textit{gensim} API\footnote{\url{https://radimrehurek.com/gensim/models/word2vec.html}}\cite{Rehurek:10}. We also compare our techniques with the BOW approach in a separate experiment using the same classifiers and the original datasets (i.e., words). In addition to the presented models explored in the document classification task, we include variations of our techniques under a different scope to evaluate their characteristics more carefully (Section~\ref{ssec:lc-behavior-results}). %TRV

%other deep learning approaches
Since one of the main objectives in this paper is to evaluate pre-trained word embeddings models, Deep Learning (DL) techniques such as BERT \cite{Devlin:18} and  XLNet \cite{Yang:19} are not readily applicable. Additionally,  the datasets in our task are considerably smaller when compared with those used in DL. Transformer-based architectures \cite{Vaswani:17} require many epochs in their fine-tuning phase and significant processing time in dedicated hardware (i.e, GPU), which make them costly restrictive. Therefore, we did not investigate such methods in our experiments but are working towards their inclusion in the future. %TRV

% SETUP
\subsection{Experiment configurations}\label{ssec:experiment_config}
%We divide our experiments into two distinct perspectives, the first for document classification (Section \ref{ssec:doc-cla-results}) and the second for lexical chains behavior analysis (Section \ref{ssec:lc-behavior-results}). 

In Section \ref{ssec:doc-cla-results}, we present the results for all variations of our models against a traditional BOW approach. In this scenario, we can analyze the different chunk sizes for our lexical chains and how the varieties of MSSA \citep{Ruas:19} affect our constructed lexical chains. This evaluation also supports the choice of our models' configurations for the comparison against the state-of-the-art systems. Thus, we compare our techniques (FL-1R, FX2-1R) with different word embeddings approaches that also use transfer learning. In Section \ref{ssec:lc-behavior-results}, we provide a more in-depth assessment of how our techniques behavior. The idea is to offer two different perspectives: (a) chunk size for the FXLC II variations and (b) the effects of MSSA models (-1R and -2R) in both FLLC II and FXLC II. %TRV

\begin{figure}
    \centering
    % \resizebox{\textwidth}{!}{
    \includegraphics[width=0.85\textwidth]{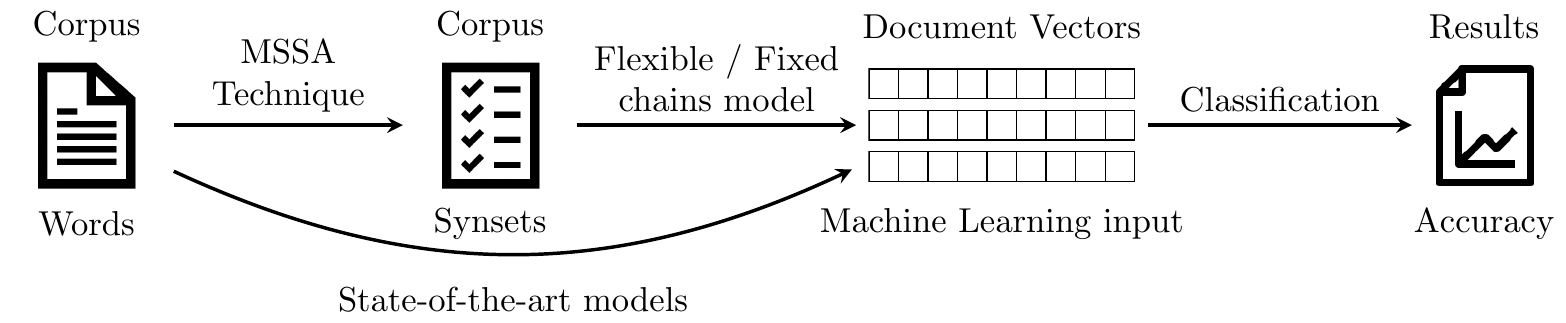}
    % }
    \caption{Pipeline for the document classification process in our experiments.}
    \label{fig:text_classification_workflow}
\end{figure}

%C data preprocessing
%All proposed models are evaluated considering the document classification task for the datasets presented in Section \ref{ssec:dataset}.
%CF Reviwer 3 (2)
The same preprocessing steps are applied to all datasets and techniques so we can guarantee more consistency in our experiments. For preprocessing, we lower-case all words in the document, remove all common English stop-words and punctuation. We represent each document as the average of its constituent word vectors using a pre-trained embeddings model. Averaging the word vectors of a document to obtain its representation is a common strategy adopted in the document classification task~\citep{Le:14, Sinoara:19}. Next, we feed them to several machine learning classifiers and evaluate their F1-Micro score, as shown in Figure~\ref{fig:text_classification_workflow}. %trv

%Figure~\ref{fig:text_classification_workflow} presents a high-level perspective of the entire process during the classification task used in our experiments.
%For each document in a corpus, we retrieve its word vectors from a pre-trained word embeddings model, average them, and build a document vector.

While the compared systems derive vectors directly from the words in each dataset, our proposed techniques require a disambiguated synset-annotated corpus. Thus, an additional step before the machine learning classifiers is necessary. We disambiguate the words from each document to obtain their respective synsets, using the MSSA~\citep{Ruas:19} algorithm. Next, as in all the compared systems, we derive the synset vectors using the pre-trained lexical chain embedding models (FLLC II and FXLC II) created from the Wikipedia dump. %TRV

We compare our findings against classical and state-of-the-art word embedding techniques that consider transfer knowledge. For traditional word embeddings techniques, we use LDA~\citep{Blei:03}, word2vec~\citep{Mikolov_a:13} and GloVe~\citep{Penni:14}. For the state-of-the-art word embedding techniques, we include fastText~\citep{Bojanowski:17}, ELMo~\citep{Matthew:18}, and USE~\citep{Cer:18}. Furthermore, we also provide a comparison of our results against a traditional BOW technique, which relies only on statistical information of the training corpus. Even though BOW does not necessarily use transfer learning from a prior task, it is probably the most used technique to represent a collection of documents in NLP. Also, the curse of dimensionality affects BOW because of the number of words in the considered vocabulary. In this manner, depending on the size of the dataset, its adoption might not be suitable. To mitigate the dimensionality problem and allow a fair comparison among the compared systems, we created a BOW representation considering the top 300 features (words), ordered by term frequency, and applying tf-idf as its weighting scheme~\citep{scikit-learn}. The only two models that do not follow this vector-length representation are ELMo and USE, with 1024 and 512 dimensions, respectively. %TRV

%document classification task
\subsection{Document classification task results} \label{ssec:doc-cla-results}
%initial text
Tables~\ref{tab:results_bow} and \ref{tab:classification_results} present the results of our experiments considering the document classification task. The results are organized as follows. Each block illustrates the results of applying all classifiers on a specific dataset. Each column represents a different word embeddings model benchmark to compare against our techniques (last two columns). Values in \bo{bold} represent the best results in a row, and \un{underlined} values the best results in a column for a specific dataset.

As explained in Section~\ref{sec:lexicalchains}, our proposed techniques are built considering a synset-annotated corpus obtained through the MSSA algorithm~\citep{Ruas:19}. For the FXLC II algorithm (Section \ref{ssec:fxlc}), we also examined the sizes of 2, 4, and 8 for the number of synsets in each chain ($chunk\_size$). Because of space limitations in Table~\ref{tab:results_bow}, the results for the chunk size of 8 for FXLC II were not included. However, we analyze their behavior for all chunk sizes in Section~\ref{ssec:lc-behavior-results}. To evaluate the best configuration for the proposed techniques, we first compared them with a traditional BOW with a tf-idf weighting scheme and 300 dimensions. We decided to keep 300 dimensions to maintain all models under similar constraints. %TRV

%In one variation of MSSA, called MSSA-NR, it is possible to control the number of times a produced synset embeddings model is used to refine the disambiguation step. Thus, we can generate a more robust semantic representation that can be used to train a new enhanced synset embeddings model. In MSSA-NR, we considered  $0\leq N\leq2$, where $N$ represents how many iterations the recurrent process is performed.

As Table~\ref{tab:results_bow} shows, BOW sustains good results for ScyGenes, and a few for ScyClusters. This behavior seems reasonable since ScyGenes and ScyClusters are the smallest datasets considered. These datasets might contain unique keywords among their categories, resulting in a reasonable classification for the BOW approach. On the other hand, as the number of documents increases, our semantic embedding representations start to overcome BOW. We believe this is due to the number of words in the corpora, as BOW is unable to extract features that lead to better classification in ambiguous and large size datasets. %TRV

\input{tables/table_results_bow.tex}

Except for the ScyClusters and ScyGenes datasets, our algorithms seem to be stable within their different variations for the same classifiers. Even though BOW achieves the best results for ScyClusters and ScyGenes, our techniques are still comparable. However, FLLC II and FXLC II do not require dedicated training in these datasets. Specifically for ScyClusters, our F1-Micro for FX2-2R deviates less than 1\% from BOW. If we consider each row in Table \ref{tab:results_bow}, FLLC II and FXLC II together present a better score in 23 out of 30 cases. We can observe that FXLC with a chunk size of 2 and FLLC II built over MSSA-1R had the majority of best results (FX2-1R and FL-1R). Thus, we used FX2-1R and FL-1R to compare against the other state-of-the-art models. %TRV

In Table~\ref{tab:classification_results}, we present the best results achieved by our Chains2Vec embedding model against state-of-the-art techniques using word embeddings. Considering the Ohsumed dataset, our approaches outperform all the other methods for all classifier baselines, which have a considerably inferior result. The Ohsumed text collection represents a difficult real-world scenario because of its size, the number of classes, and constituent words. In this dataset, our fixed lexical chains of size 2 (FX2-1R) achieves the best results for all classifiers. For the 20Newsgroups dataset, our techniques overcome the baselines for the $K$-NN, RF and NB, while fastText was superior considering LR and SVM. %TRV %constituent words \citep{Yang:99}

\input{tables/table_results.tex}

Considering the Reuters-21578 dataset, GloVe achieves the best results in 4 out of 5 classifiers. If we think how GloVe uses a co-occurrence approach to build its vectors, this outcome is expected. Several documents in this dataset are composed of short phrases, which prevents our techniques from deriving a good semantic representation. As a consequence, our chains performed poorly. However, even affected by Reuters' content, we outperform methods able to embed any given token, such as ELMo and USE. On BBC, the synset embedding models present the highest F1-Micro score in 4 of the 5 classifiers, in which our lexical chain representation has the best scores for RF and LR. The documents in the BBC dataset are composed of BBC News articles, which use a formal and cohesive English. A coherent text structure contributes to the semantic representation in our techniques and M1R since they are based on WordNet.  %TRV

%pre-trained models vs ELMo and USE + classifiers using DL 
When we compare the results of word2vec, GloVe, and fastText with USE and ELMo we see the former group outperforms the latter in almost all cases in their best configuration (except for GloVe and ELMo in Ohsumed). We consider two reasons for this behavior, the difference in their base training corpus and the lack of fine-tuning in the neural network inspired models. While GloVe and fastText mainly use Wikipedia dumps as their training corpus, ELMo and USE are trained in the 1 Billion Word Benchmark and various sources, as described in Table \ref{tab:model_details}. All pre-trained word embeddings models in our experiments were not fine-tuned since the main goal was to compare their behavior when applied directly to our task. As a result, ELMo and USE showed a significant disadvantage when compared to the other models since they were using the frozen weights from their general base training tasks. In the future, we plan to explore the benefits of fine-tuning neural network models (e.g., Bi-LSTM, Transformers) towards a specific task by comparing them with our techniques and incorporating the most prominent ones to the best performing neural architectures. %TRV 

Finally, on ScyClusters and ScyGenes datasets, our lexical chains show superior results in 7 out of 10 experiments. If we consider word2vec and FX2-1R, for the $K$-NN classifier, our technique shows an improvement of 15\% and 9\% for the ScyClusters and ScyGenes datasets respectively. These results suggest that the semantic relations extracted by our algorithms improve the quality of a standard word embeddings. As in BBC and Oshumed, ScyClusters and ScyGenes contain formal, cohesive, and typo-free English documents. In an overall analysis, we sustain good results throughout the explored datasets. The synset techniques, especially the proposed ones, are able to extract semantic relations within the documents properly. Besides, the generated synset embeddings models are also significantly smaller (75\%) than the ones compared in this paper. %TRV

% Lexical chains behavior
\subsection{Lexical chains behavior analysis} \label{ssec:lc-behavior-results}
\begin{figure}[t!]
    \centering
    
    \setcounter{subfigure}{0}% reset the counter 
    \subfigure[Ohsumed.]{\includegraphics[scale=0.55]{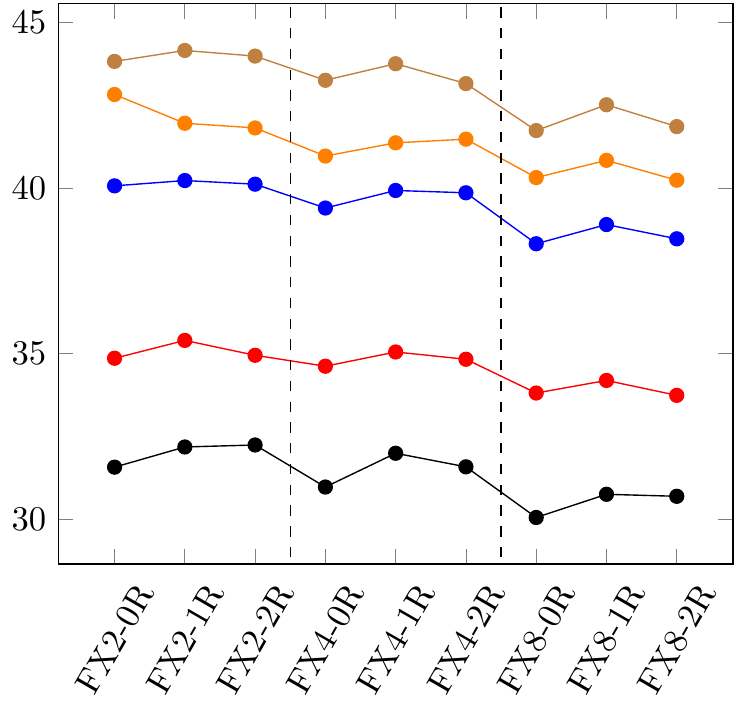}}
    \subfigure[20Newsgroups.]{\includegraphics[scale=0.55]{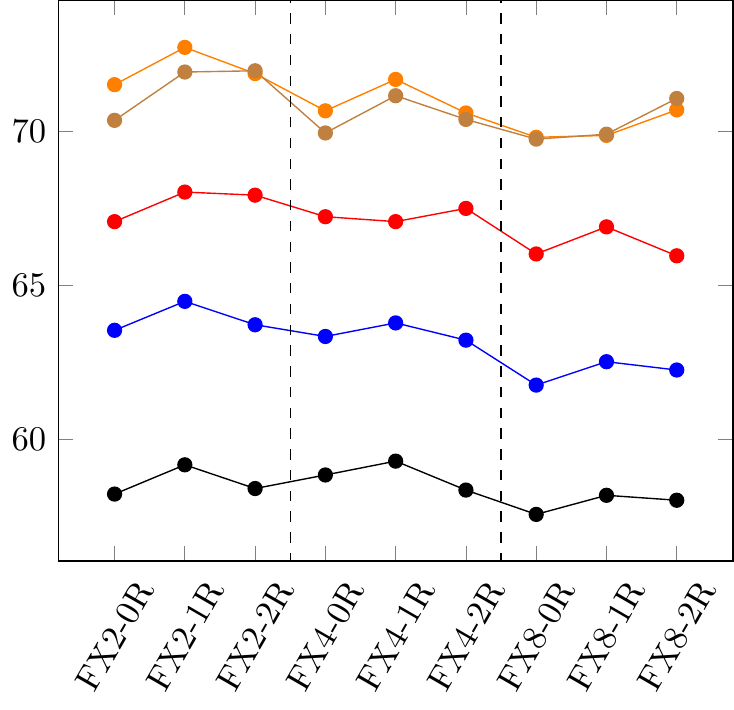}}
    \subfigure[Reuters-21578.]{\includegraphics[scale=0.55]{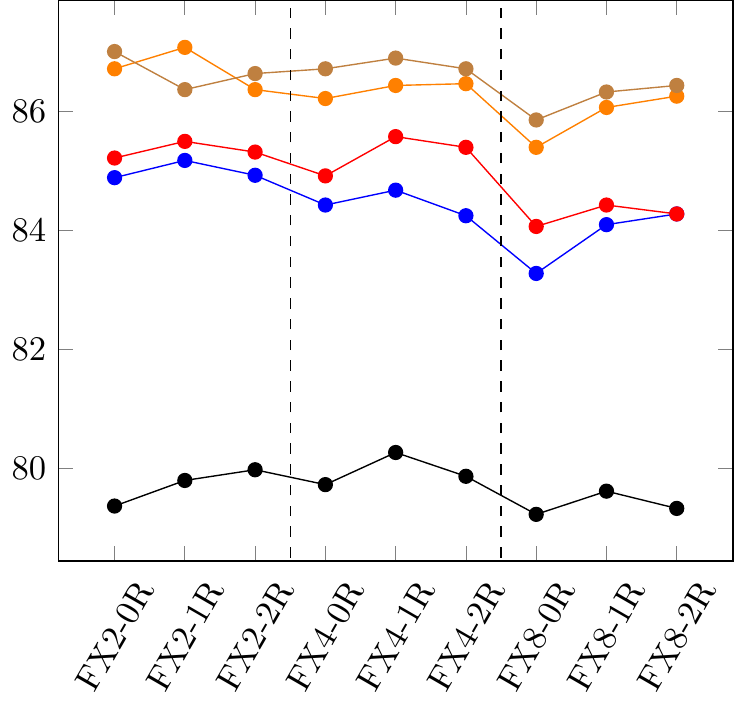}}
    \subfigure[BBC.]{\includegraphics[scale=0.55]{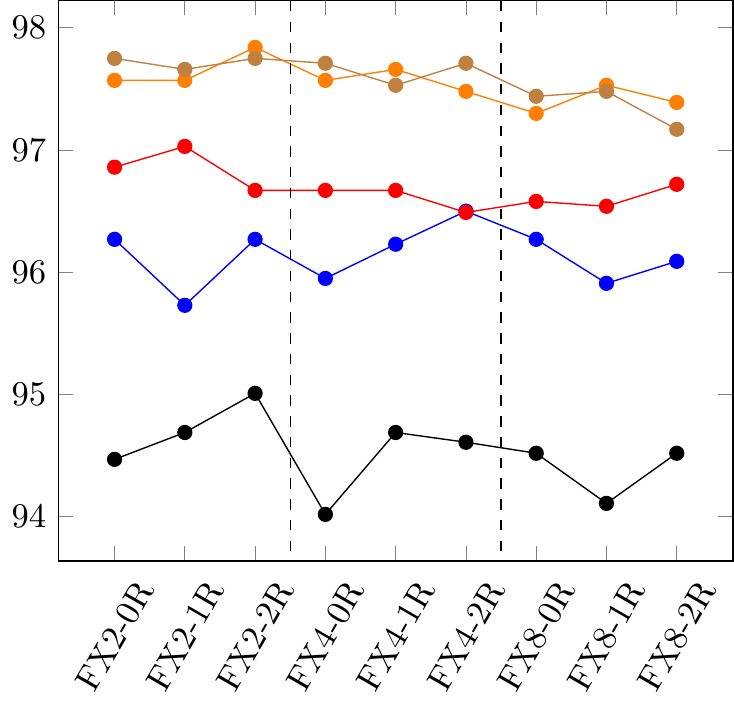}}
    \subfigure[ScyClusters.]{\includegraphics[scale=0.55]{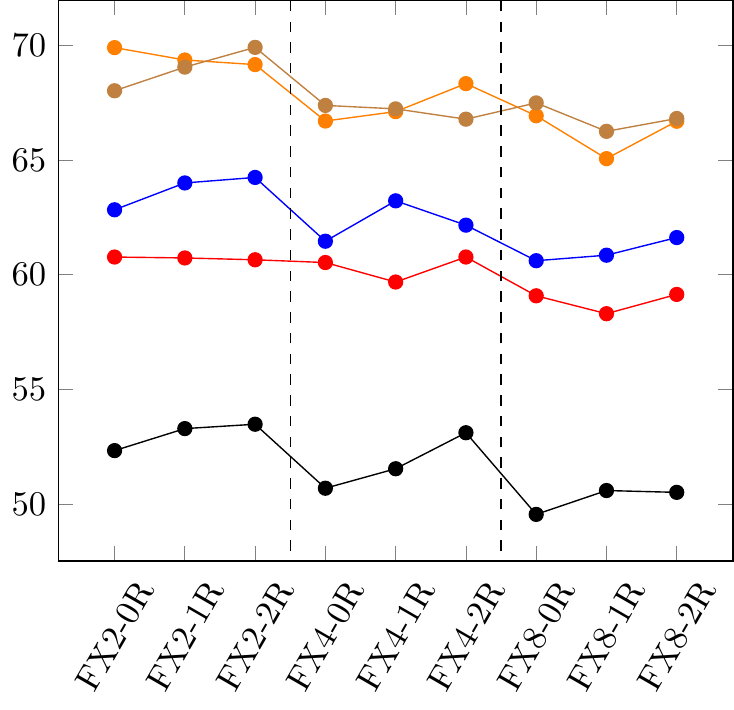}}
    \subfigure[ScyGenes.]{\includegraphics[scale=0.55]{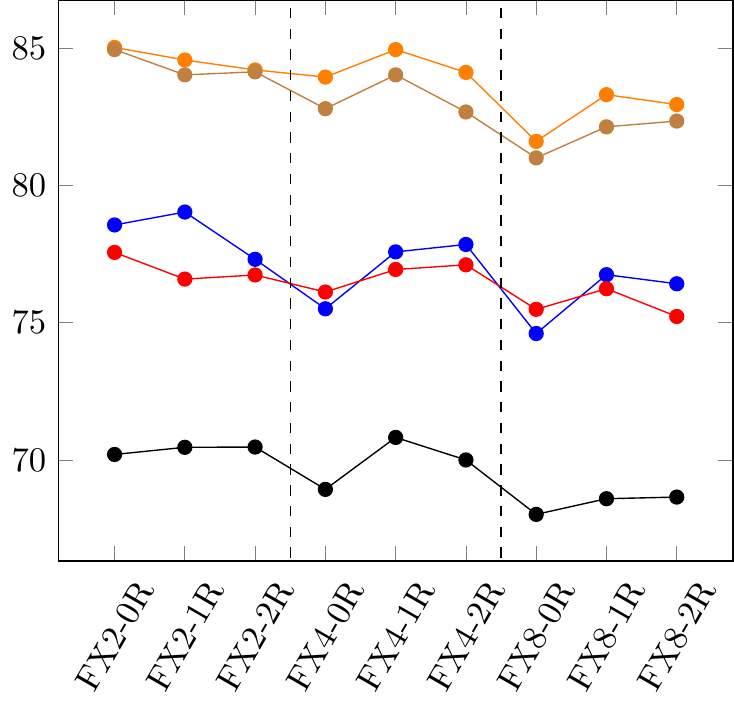}}
    \subfigure{\includegraphics[scale=0.85]{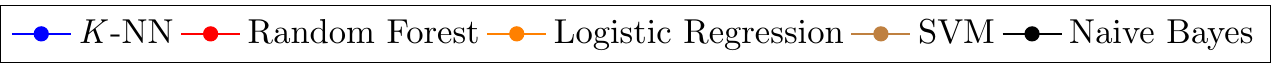}}
    \caption{F1-Micro score for FXLC II with variable chunk size and pre-trained MSSA models.} \label{fx:behavior}
\end{figure}

In this section, we investigate how our proposed techniques are affected by their internal configurations. We still keep the same machine learning classifiers and datasets to maintain consistency in our comparisons. The main idea is to provide a different perspective on how the fixed chains are affected by the chosen chunk size and the pre-trained models in which they are based (MSSA-1R and MSSA-2R). The results in Figure~\ref{fx:behavior} show the F1-Micro score over different variations of our approaches, for six datasets and five classifiers. %TRV

%R1(2)
The idea behind both techniques (FLLC and FXLC) is to provide a more concise and robust document representation for text documents. Techniques such as word2vec and GloVe have two significant limitations: (i) they represent all word senses (e.g., \textit{java-coffee}, \textit{java-island}, \textit{java-programming-language}) in a single vector and (ii) ignore the relationship between words; both of which our proposed techniques do. While MSSA solves the former (i), FLLC and FXLC focus on the latter (ii). MSSA fixes the multi-sense representation, but the relationship between word works at a local level. On the other hand, FLLC and FXLC work on the relationship between consecutive words of the entire document at a multi-sense level. Regardless of how many synsets each chain in FLLC and FXLC has, we represent them with just one. Thus, reducing the necessary features for the document classification task and still keeping all benefits of the word embeddings algorithms in which they were trained.  %TRV

For FXLC II in Figure~\ref{fx:behavior}, all models built on top of MSSA-1R (1R) present an improvement if compared with their base version (0R). However, this improvement is not valid for the -2R variant. Our fixed chains seem to be more affected by the change in their chunk-size than the pre-trained model used (-1R and -2R). If we consider each block separately (vertical dashed lines), the F1-Micro score for each fixed chain size remains quite stable. Nevertheless, moving from 2 to 8 synsets per chain affects our technique significantly. In other words, the chunk size parameter in our algorithm is inversely proportional to the quality of semantic representation in our chains. Thus, resulting in a decrease in the score when we move from 2 to 8 synsets per chain. For a chunk size of 1, FXLC is reduced to MSSA, since each word-sense is its own chain. Since the lexical chains produced with the FLLC II algorithm present the same behavior as in the FXLC, concerning the pre-trained MSSA model used, but on a much smaller variation scale, we decided not to include its graphical analysis. %TRV

%We also provide the same perspective analysis considering the FLLC II technique in Table \ref{fl:behavior}. Since the only variation in the flexible chains is the pre-trained model used to derive its chains, we have a smaller scenario if compared to the FXLC II alternative. Another common aspect to both techniques is the stability regarding the machine learning classifiers. For both, FLLC II and FXLC II, the best results are always between logistic regression and SVM, independently of the dataset, change of chunk size, and recurrence variation.

%TR - include possible characteristic from LR and SVM that supports our model

%\begin{figure}[t!]
%    \centering
%    \subfigure{\includegraphics[scale=0.95]{images/graphs/legend_crop.pdf}}
%    \setcounter{subfigure}{0} % reset the counter
%    \subfigure[Ohsumed.]{\includegraphics[scale=0.53]{images/graphs/ohsumed_fllc.pdf}}
%    \subfigure[20newsgroups.]{\includegraphics[scale=0.53]{images/graphs/20news_fllc.pdf}}
%    \subfigure[Reuters-21578.]{\includegraphics[scale=0.53]{images/graphs/reuters_fllc.pdf}}
%    \subfigure[BBC.]{\includegraphics[scale=0.53]{images/graphs/bbc_fllc.pdf}}
%    \subfigure[ScyClusters.]{\includegraphics[scale=0.53]{images/graphs/scycluster_fllc.pdf}}
%    \subfigure[ScyGenes.]{\includegraphics[scale=0.53]{images/graphs/scygenes_fllc.pdf}}
 %   \caption{F1-Micro score for flexible lexical chains using variations of pre-trained MSSA models.}  \label{fl:behavior}
%\end{figure}

%% file: tables/table_datasets.tex
\begin{table}[htb]
\centering
  \caption{Technical details about the datasets after preprocessing.}
  \label{tab:dataset_description}
  \resizebox{0.93\textwidth}{!}{ 
    \begin{tabular}{llrrrr}
      \toprule
      Corpus                                        & Subject               & \#docs & \#classes & \#tokens & \#synsets  \\\midrule
      Ohsumed~\citep{Joachims:98a}                 & Medical abstracts     & 56984  &  23       & 64154    & 36395      \\
      20Newsgroups~\citep{Joachims:98a}             & News                  & 18846  &  20       & 129782   & 43413      \\
      Reuters-21578                                 & News                  & 9980   &  10       & 24273    & 21747      \\
      BBC~\citep{Greene:06}                         & News                  & 2225   &  5        & 29126    & 29151      \\
      ScyClusters~\citep{Medeiros:05}   & Biological abstracts  & 1655   &  7        & 13265    & 11428      \\
      ScyGenes~\citep{Medeiros:05}      & Biological abstracts  & 1114   &  7        & 10553    & 10045      \\
      
      \bottomrule
    \end{tabular}
  }
\end{table}

%% file: tables/table_modeldetails.tex
\begin{table}
\centering
\caption{Word embeddings used and their main characteristics. * For USE, \cite{Cer:18} reports its training data as a collection of sources from Wikipedia, web news, web question-answer pages discussion forums and Stanford Natural Language Inference (SNLI) corpus.}     \label{tab:model_details}
\resizebox{0.99\textwidth}{!}{
\begin{tabular}{lllr}\toprule
Algorithm & Main Characteristics                      & Training Corpus                     & Dimensions  \\ \midrule
LDA       & Probability distribution                  & Wikipedia Dump 2010                 & 300         \\
word2vec  & Continuous Bag-of-Words (CBOW)                   & Google News                         & 300         \\
GloVe     & Word-word co-occurrence matrix            & Wikipedia dump 2014 + Gigaword 5    & 300         \\
fastText  & Skip-gram                                  & Wikipedia Dump 2017 + UMBC          & 300         \\
USE       & Deep Average Network                      & Various sources*                    & 512         \\
ELMo      & Bidirectional Long Short Term Memory      & 1 Billion Word Benchmark            & 1024        \\
MSSA-1R       & Most Suitable Sense Annotation -1R         & Wikipedia Dump 2010                 & 300         \\
FL-1R     & Flexible Lexical Chains II + MSSA-1R                & Wikipedia Dump 2010           & 300         \\
FX2-1R   & Fixed Lexical Chains II + MSSA-1R                   & Wikipedia Dump 2010           & 300         \\  \bottomrule
\end{tabular}
}
\end{table}

%% file: tables/table_results_bow.tex
\begin{table}[!t]
    \caption{Classification F1-Micro score for BOW approach against the proposed techniques for each classifier and dataset. Values in \textbf{bold} represent the best result of that row. \underline{Underline} values represent the best value in that column.}% $K$-NN - K Nearest Neighbors; RF - Random Forest; LR - Logistic Regression; SVM - Support Vector Machine; NB - Naive Bayes.} 
    \label{tab:results_bow}
    \centering
   \resizebox{0.97\textwidth}{!}{ 
    \begin{tabular}{lrrrrrrrrrrrrr}
    \toprule
           & BOW            & FL-0R        & FL-1R        & FL-2R         & FX2-0R      & FX2-1R      & FX2-2R      & FX4-0R        & FX4-1R      & FX4-2R  \\\midrule %& FX8-0R & FX8-1R & FX8-2R\\\midrule
Ohsumed    & 0.3486         & 0.4402       & 0.4412       & 0.4389        & 0.4383      & \un{0.4416} & 0.4399      & 0.4326        & 0.4376      & 0.4316  \\\midrule %& 0.4174 & 0.4252 & 0.4186 \\\midrule
$K$-NN     & 0.3187         & 0.3953       & 0.3967       & 0.3936        & 0.4007      & \bo{0.4023} & 0.4012      & 0.3940        & 0.3993      & 0.3986  \\         %& 0.3832 & 0.3890 & 0.3847 \\
RF         & 0.3486         & 0.3407       & 0.3397       & 0.3380        & 0.3486      & \bo{0.3540} & 0.3495      & 0.3462        & 0.3505      & 0.3483  \\         %& 0.3381 & 0.3419 & 0.3374 \\
LR         & 0.3441         & 0.4123       & 0.4169       & 0.4151        & \bo{0.4283} & 0.4196      & 0.4182      & 0.4097        & 0.4137      & 0.4148  \\         %& 0.4032 & 0.4084 & 0.4024 \\
SVM        & 0.3364         & 0.4402       & 0.4412       & 0.4389        & 0.4383      & \bo{0.4416} & 0.4399      & 0.4326        & 0.4376      & 0.4316  \\         %& 0.4174 & 0.4252 & 0.4186 \\
NB         & 0.1662         & 0.3091       & 0.3065       & 0.3079        & 0.3157      & 0.3218      & \bo{0.3224} & 0.3097        & 0.3199      & 0.3158  \\\midrule %& 0.3005 & 0.3075 & 0.3069 \\\midrule

20News     & 0.5458         & 0.7135       & 0.7147       & 0.7167        & 0.7151      & \un{0.7272} & 0.7196      & 0.7066        & 0.7168      &  0.7059  \\\midrule %& 0.6980  & 0.6990  & 0.7106 \\\midrule
$K$-NN     & 0.4316         & 0.6312       & 0.6329       & 0.6298        & 0.6353      & \bo{0.6447} & 0.6371      & 0.6333        & 0.6377      &  0.6321  \\         %& 0.6175  & 0.6251  & 0.6224 \\
RF         & 0.5458         & 0.6701       & 0.6742       & 0.6642        & 0.6706      & \bo{0.6802} & 0.6792      & 0.6722        & 0.6706      &  0.6749  \\         %& 0.6601  & 0.6689  & 0.6595 \\
LR         & 0.5252         & 0.7067       & 0.7147       & 0.7131        & 0.7151      & \bo{0.7272} & 0.7187      & 0.7066        & 0.7168      &  0.7059  \\         %& 0.6980  & 0.6986  & 0.7069 \\
SVM        & 0.5096         & 0.7135       & 0.7114       & 0.7167        & 0.7035      & 0.7192      & \bo{0.7196} & 0.6994        & 0.7115      &  0.7038  \\         %& 0.6974  & 0.6990  & 0.7106 \\
NB         & 0.3946         & 0.5860       & 0.5884       & 0.5890        & 0.5821      & 0.5916      & 0.5839      & 0.5883        & \bo{0.5928} &  0.5834  \\\midrule %& 0.5755  & 0.5817  & 0.5801 \\\midrule

Reuters    & 0.8683         & 0.8719       & \un{0.8726}  & 0.8719        & 0.8701      & 0.8708      & 0.8664      & 0.8672        & 0.8690       &  0.8672  \\\midrule %& 0.8586  & 0.8633  & 0.8644 \\\midrule
$K$-NN     & 0.8378         & \bo{0.8558}  & 0.8554       & 0.8539        & 0.8489      & 0.8518      & 0.8493      & 0.8443        & 0.8468       &  0.8425  \\         %& 0.8328  & 0.8410  & 0.8428 \\
RF         & 0.8561         & 0.8543       & 0.8525       & 0.8504        & 0.8522      & 0.8550      & 0.8532      & 0.8492        & \bo{0.8558}  &  0.8540  \\         %& 0.8407  & 0.8443  & 0.8428 \\
LR         & 0.8683         & 0.8698       & \bo{0.8726}  & 0.8676        & 0.8672      & 0.8708      & 0.8637      & 0.8622        & 0.8644       &  0.8647  \\         %& 0.8540  & 0.8607  & 0.8626 \\
SVM        & 0.6232         & \bo{0.8719}  & 0.8640       & \bo{0.8719}   & 0.8701      & 0.8637      & 0.8664      & 0.8672        & 0.8690       &  0.8672  \\         %& 0.8586  & 0.8633  & 0.8644 \\
NB         & 0.7718         & 0.8116       & \bo{0.8120}  & 0.8044        & 0.7937      & 0.7980      & 0.7998      & 0.7973        & 0.8027       &  0.7987  \\\midrule %& 0.7923  & 0.7962  & 0.7933 \\\midrule

BBC        & 0.9524         & \un{0.9784}  & \un{0.9784}  & 0.9771        & 0.9775      & 0.9757      & \un{0.9784} & 0.9771        & 0.9766       & 0.9771      \\\midrule %& 0.9744  & 0.9753  & 0.9739 \\\midrule
$K$-NN     & 0.9097         & 0.9604       & 0.9600       & 0.9627        & 0.9627      & 0.9573      & 0.9627      & 0.9595        & 0.9623       & \bo{0.9650} \\         %& 0.9627  & 0.9591  & 0.9609 \\
RF         & 0.9421         & 0.9645       & 0.9667       & 0.9636        & 0.9686      & \bo{0.9703} & 0.9667      & 0.9667        & 0.9667       & 0.9649      \\         %& 0.9658  & 0.9654  & 0.9672 \\
LR         & 0.9524         & \bo{0.9784}  & \bo{0.9784}  & 0.9753        & 0.9757      & 0.9757      & \bo{0.9784} & 0.9757        & 0.9766       & 0.9748      \\         %& 0.9730  & 0.9753  & 0.9739 \\
SVM        & 0.9510         & \bo{0.9780}  & 0.9771       & 0.9771        & 0.9775      & 0.9766      & 0.9775      & 0.9771        & 0.9753       & 0.9771      \\         %& 0.9744  & 0.9748  & 0.9717 \\
NB         & 0.9137         & 0.9474       & 0.9456       & 0.9465        & 0.9447      & 0.9469      & \bo{0.9501} & 0.9402        & 0.9469       & 0.9461      \\\midrule %& 0.9452  & 0.9411  & 0.9452 \\\midrule

ScyClusters  & \un{0.6997}  & 0.6930       & 0.6822       & 0.6827        & 0.6990      & 0.6936      & 0.6991      & 0.6738        & 0.6723       & 0.6833      \\\midrule %& 0.6749  & 0.6625  & 0.6669 \\\midrule
$K$-NN     & \bo{0.6470}    & 0.6430       & 0.6165       & 0.6340        & 0.6283      & 0.6400      & 0.6424      & 0.6146        & 0.6322       & 0.6216      \\         %& 0.6061  & 0.6085  & 0.6162 \\
RF         & \bo{0.6997}    & 0.5975       & 0.5940       & 0.5885        & 0.6077      & 0.6073      & 0.6065      & 0.6053        & 0.5968       & 0.6077      \\         %& 0.5908  & 0.5830  & 0.5914 \\
LR         & 0.6616         & 0.6845       & 0.6822       & 0.6827        & \bo{0.6990} & 0.6936      & 0.6916      & 0.6670        & 0.6711       & 0.6833      \\         %& 0.6693  & 0.6506  & 0.6669 \\
SVM        & 0.6610         & 0.6930       & 0.6701       & 0.6815        & 0.6802      & 0.6905      & \bo{0.6991} & 0.6738        & 0.6723       & 0.6678      \\         %& 0.6749  & 0.6625  & 0.6681 \\
NB         & 0.5329         & 0.5354       & \bo{0.5385}  & 0.5324        & 0.5233      & 0.5329      & 0.5348      & 0.5069        & 0.5154       & 0.5311      \\\midrule %& 0.4955  & 0.5059  & 0.5051 \\\midrule

ScyGenes   & \un{0.9767}    & 0.8474       & 0.8521       & 0.8457        & 0.8502      & 0.8456      & 0.8420      & 0.8394        & 0.8494       & 0.8411      \\\midrule %&  0.8160 & 0.8330  & 0.8294 \\\midrule
$K$-NN     & \bo{0.9094}    & 0.7829       & 0.7750       & 0.7783        & 0.7856      & 0.7903      & 0.7731      & 0.7551        & 0.7758       & 0.7785      \\         %&  0.7461 & 0.7675  & 0.7642 \\
RF         & \bo{0.9767}    & 0.7727       & 0.7603       & 0.7597        & 0.7756      & 0.7659      & 0.7674      & 0.7612        & 0.7694       & 0.7711      \\         %&  0.7549 & 0.7624  & 0.7523 \\
LR         & \bo{0.9309}    & 0.8454       & 0.8521       & 0.8457        & 0.8502      & 0.8456      & 0.8420      & 0.8394        & 0.8494       & 0.8411      \\         %&  0.8160 & 0.8330  & 0.8294 \\
SVM        & \bo{0.9282}    & 0.8474       & 0.8475       & 0.8450        & 0.8494      & 0.8402      & 0.8413      & 0.8279        & 0.8402       & 0.8267      \\         %&  0.8100 & 0.8213  & 0.8234 \\
NB         & \bo{0.9229}    & 0.6858       & 0.6895       & 0.6967        & 0.7021      & 0.7047      & 0.7048      & 0.6894        & 0.7083       & 0.7001      \\         %&  0.6803 & 0.6860  & 0.6866 \\
\bottomrule
    \end{tabular}
    }
\end{table}

%% file: tables/table_results.tex
\begin{table}[!t]
\centering
\caption{Classification F1-Micro score for word embeddings models against proposed techniques for each classifier and dataset. Values in \textbf{bold} represent the best result of that row. \underline{Underline} values represent the best value in that column.} %$K$-NN - K Nearest Neighbors; RF - Random Forest; LR - Logistic Regression; SVM - Support Vector Machine; NB - Naive Bayes.}     
\label{tab:classification_results}
\resizebox{0.90\textwidth}{!}{
\begin{tabular}{lrrrrrrrrr}\toprule
                & LDA      & w2v      &  GloVe         & fastText      & USE      & ELMo           &  MSSA-1R             & FL-1R        & FX2-1R      \\\midrule

Ohsumed         & 0.2262   & 0.4223        &  0.4136        & 0.4324        & 0.3009   & 0.4172         &  0.4357          & 0.4412        & \un{0.4416}  \\ \midrule
$K$-NN          & 0.2138   & 0.3822        & 0.3731         & 0.3912        & 0.3009   &  0.3209        &  0.3975          & 0.3961        & \bo{0.4023}  \\ 
RF              & 0.1689   & 0.3302        & 0.3258         & 0.3447        & 0.2899   &  0.2851        &  0.3411          & 0.3397        & \bo{0.3540}  \\ 
LR              & 0.2262   & 0.3981        & 0.4136         & 0.4171        & 0.2792   &  0.4172        &  0.4185          & 0.4169        & \bo{0.4196}  \\ 
SVM             & 0.2056   & 0.4223        & 0.3286         & 0.4324        & 0.2542   &  0.3194        &  0.4357          & 0.4412        & \bo{0.4416}  \\ 
NB              & 0.0558   & 0.2820        & 0.2771         & 0.2785        & 0.1947   &  0.1865        &  0.3118          & 0.3065        & \bo{0.3218}  \\\midrule

20News    & 0.6340   & 0.7110        &  0.7153        & \un{0.7485}   & 0.6476   & 0.6895         &  0.7201          & 0.7147        & 0.7272       \\ \midrule
$K$-NN          & 0.5013   & 0.5737        & 0.5425         & 0.6134        & 0.5588   &  0.4574        &  0.6320          & 0.6329        & \bo{0.6447}  \\
RF              & 0.6340   & 0.6300        & 0.6386         & 0.6794        & 0.6476   &  0.5389        &  0.6753          & 0.6742        & \bo{0.6802}  \\
LR              & 0.5498   & 0.6657        & 0.7152         & \bo{0.7288}   & 0.5447   &  0.6895        &  0.7167          & 0.7147        & 0.7272       \\ 
SVM             & 0.5388   & 0.7110        & 0.7153         & \bo{0.7485}   & 0.5171   &  0.4981        &  0.7201          & 0.7114        & 0.7192       \\
NB              & 0.4624   & 0.4426        & 0.4133         & 0.5104        & 0.4401   &  0.2677        &  0.5895          & 0.5884        & \bo{0.5916}  \\\midrule

Reuters   & 0.8260   & 0.8805        & \un{0.8830}    & 0.8802        & 0.8278   & 0.8780         &  0.8719          & 0.8726        & 0.8708       \\ \midrule
$K$-NN          & 0.7919   & \bo{0.8680}   & 0.8640         & 0.8593        & 0.8267   &  0.8436        &  0.8568          & 0.8554        & 0.8518       \\
RF              & 0.8260   & 0.8561        & \bo{0.8619}    & 0.8536        & 0.8278   &  0.8281        &  0.8550          & 0.8525        & 0.8550       \\ 
LR              & 0.7875   & 0.8698        & \bo{0.8776}    & 0.8705        & 0.7370   &  0.8751        &  0.8651          & 0.8726        & 0.8708       \\ 
SVM             & 0.8041   & 0.8805        & \bo{0.8830}    & 0.8802        & 0.7951   &  0.8780        &  0.8719          & 0.8640        & 0.8637       \\
NB              & 0.6024   & 0.7740        & \bo{0.8224}    & 0.8034        & 0.7725   &  0.7603        &  0.8009          & 0.8120        & 0.7980       \\\midrule

BBC             & 0.9552   & 0.9708        & \un{0.9784}    & 0.9766        & 0.9672   & 0.9743         &  0.9780          & \un{0.9784}   & 0.9766       \\\midrule
$K$-NN          & 0.9304   & 0.9591        & \bo{0.9622}    & 0.9618        & 0.9577   & 0.9497         &  0.9596          & 0.9600        & 0.9573       \\
RF              & 0.9552   & 0.9532        & 0.9631         & 0.9663        & 0.9672   & 0.9573         &  0.9681          & 0.9667        & \bo{0.9703}  \\ 
LR              & 0.9241   & 0.9478        & 0.9690         & 0.9681        & 0.9370   & 0.9649         &  0.9766          & \bo{0.9784}   & 0.9757       \\
SVM             & 0.9295   & 0.9708        & \bo{0.9784}    & 0.9766        & 0.9474   & 0.9743         &  0.9780          & 0.9771        & 0.9766       \\
NB              & 0.8680   & 0.9218        & 0.9483         & 0.9469        & 0.9442   & 0.9195         &  \bo{0.9501}     & 0.9456        & 0.9469       \\\midrule

ScyClusters     & 0.4814   & 0.6410        & 0.6391         & 0.6645        & 0.4966   & 0.6612         &  \un{0.7027}     & 0.6822        & 0.6936       \\ \midrule
$K$-NN          & 0.4137   & 0.5919        & 0.5903         & 0.5777        & 0.4835   &  0.5675        &  0.6267          & 0.6165        & \bo{0.6400}  \\
RF              & 0.4814   & 0.5479        & 0.5469         & 0.5890        & 0.4966   &  0.5317        &  0.5889          & 0.5940        & \bo{0.6073}  \\ 
LR              & 0.3692   & 0.5879        & 0.6168         & 0.6162        & 0.3475   &  0.6612        &  0.6839          & 0.6822        & \bo{0.6936}  \\ 
SVM             & 0.3439   & 0.6410        & 0.6391         & 0.6645        & 0.3439   &  0.3650        &  \bo{0.7027}     & 0.6701        & 0.6905       \\
NB              & 0.2362   & 0.4769        & 0.5185         & 0.4790        & 0.3910   &  0.4670        &  0.5373          & \bo{0.5385}   & 0.5329       \\ \midrule
    
ScyGenes        & 0.6104   & 0.7988        & 0.7961         & 0.8301        & 0.6460   & \un{0.8761}    &  0.8556          & 0.8521        & 0.8456       \\ \midrule
$K$-NN          & 0.4573   & 0.7085        & 0.7120         & 0.7281        & 0.6105   &  0.7480        &  0.7866          & 0.7750        & \bo{0.7903}  \\
RF              & 0.6104   & 0.6849        & 0.7363         & 0.7308        & 0.6460   &  0.7423        &  0.7649          & 0.7603        & \bo{0.7659}  \\  
LR              & 0.3488   & 0.6346        & 0.7746         & 0.7477        & 0.3044   &  0.8509        &  \bo{0.8556}     & 0.8521        & 0.8456       \\ 
SVM             & 0.3471   & 0.7988        & 0.7961         & 0.8301        & 0.2442   &  \bo{0.8761}   &  0.8538          & 0.8475        & 0.8402       \\
NB              & 0.3766   & 0.6326        & 0.6743         & 0.6249        & 0.5725   &  0.7028        &  0.6977          & 0.6895        & \bo{0.7047}  \\
\bottomrule
\end{tabular}
}
\end{table}

%% file: limitations.tex
In this section, we provide a more in-depth discussion about the main aspects of our techniques, point out their strengths, and discuss possible alternatives to mitigate some of their limitations. %TRV

%main idea + other lexical databases
The main objective of our proposed algorithms is to combine the semantic relations of the lexical chains to provide lightweight pre-trained models. These models generate synset-vectors suited to improve the predictive abilities of classifier algorithms. We represent our lexical chains through synset-tokens. For the FLLC II technique (Section~\ref{ssec:fllc}), we use WordNet to identify how the synsets in our documents are connected. As a result, we remove tokens not mapped in WordNet~\citep{Fellbaum:98}, which might not reflect the true semantic value in the document. Other lexical databases, such as ConceptNet~\citep{LiuCN:04} and BabelNet~\citep{Navigli:12}, might provide alternative structures for the synsets. However, ConceptNet does not structure its components in sets of synonyms, which would require drastic changes in our algorithm. On the other hand, BabelNet uses a similar synset structure to WordNet that could be explored by our algorithm. BabelNet integrates different resources\footnote{\url{https://babelnet.org/about}} (including WordNet). Unfortunately, because of their proprietary license, BabelNet's access is not as facilitated as ConceptNet and WordNet. It is important to mention that, for research purposes, BabelNet indices are available upon an application request to their company. The application requires affiliation with a research institution or a Ph.D. student status, besides the non-commercial nature of the project. %TRV

% why not others word embeddings
One might point out that the proposed techniques only use a word2vec implementation to embed the synset corpora produced by FLLC II and FXLC II. Although this brings an interesting perspective, we decided to validate our algorithms using a straightforward method before moving to more complex ones, such as fastText~\citep{Bojanowski:17}, ELMo~\citep{Matthew:18}, and USE~\citep{Cer:18}. In fastText, they propose to learn word representations as a sum of the $n$-grams of its constituent sub-words. Sub-words would incorporate a high number of tokens that do not exist in WordNet. Thus, our approaches would not take advantage of these extra computations. In ELMo, the sub-word issue is even stronger since their words vectors are a linear combination of their characters.  We also examined the recently published USE~\citep{Cer:18}, but their implementation only allows us to access and retrieve word vectors from their pre-calculated model, not train a new corpus. Another factor that prevents us from using ELMo\footnote{\url{https://github.com/allenai/bilm-tf}}, and any Transfomer-inspired architectures (e.g., BERT) for now, is their expensive training and fine-tuning processes. %TRV

The synsets in our chains and embeddings models are built using proper English. For that reason, our approach does not generalize that well for documents containing informal English (e.g., jargon). A way to mitigate the lack of matches between document tokens and the embeddings model would be to incorporate multiple pre-trained word embeddings, similar to what \citet{Sinoara:19} adopt. However, this can lead to an overhead as significant as the pre-trained models considered. Additionally, if several word embeddings models have the same word, a ranking system would need to be used as well. %TRV

%sysnet objects
Even though we incorporate as much as 19 semantic objects\textsuperscript{\ref{wordnet_atributte_list}} in WordNet for the FLLC II algorithm~\ref{ssec:fllc}, there are other artifacts in WordNet. Nevertheless, during the early stages of our research, we discovered that many of synset's attributes ($related\_synsets$ in Algorithm~\ref{alg:fllc}) would rarely return any synset. By decreasing the number of related synsets, we could gain more performance during the construction of our flexible chains. In addition to the number of available attributes, we did not explore deeper levels of relations for each object (e.g., hypernyms of hypernyms). In other words, for each synset in the related synsets, we did not investigate their related synsets.

%best results
During our experiments, we noticed that datasets composed of paper abstracts had the best performance. These results reinforce the strength of our technique on documents in well-written English. Even so, we still had excellent results on other datasets of other nature (e.g., news). Especially, with 20Newsgroups we had 3 out of 5 best results. During our investigations on the experiments against BOW, we could understand that our performance tends to decrease when we use chains with large chunks sizes. A possible explanation may be because with larger chain chunks, we have a stronger dimensionality reduction, and consequently, we might lose too much information in the process. %TRV

%% file: conclusions.tex
%introduction link
In this work, we proposed two techniques which combine the areas of lexical chains and word embeddings to generate a lightweight pre-trained model capable of improving the document classification. FLLC II and FXLC II help us to extract implicit semantic relations between words in a synset-based text document using different methods. For FLLC II, we build our lexical chains with the assistance of a lexical database to extract the relations among the constituent synsets of a document. Additionally, FLLC II can handle any POS and considers 19 synset objects\textsuperscript{\ref{wordnet_atributte_list}} in WordNet. In FXLC II, we pre-define a specific number of synsets for each chain. In both methods, FLLC II and FXLC II, we produce a more concise and robust semantic representation than word-based (e.g., BOW) techniques and traditional word embeddings (e.g., word2vec). %TRV %In other words, while FLLC II offers a dynamic approach in capturing the different ideas in a text, FXLC II assumes a more strict methodology and enforces a limit for each chain. 

%validation/results maybe more details on the results - link with last section with more numbers/stats
Our experiments compared the traditional BOW approach and seven other state-of-the-art techniques: LDA, word2vec, GloVe, fastText, USE, ELMo, and MSSA. To explore the stability of our systems, we evaluated all methods on six distinct datasets with specific characteristics that impose a particular challenge on each classification. Furthermore, we considered five machine learning classifiers in our experiments so we can guarantee our findings are not bound to one specific classification technique. Our results showed the proposed methods for building lexical chains leverage the semantic representation offered in previous contributions. As Table~\ref{tab:classification_results} shows, FLLC II and FXLC II often improve the results of MSSA and traditional word2vec, which are their building blocks. Therefore, we believe the proposed techniques applicable to other NLP downstream tasks that also require a refined semantic representation, such as sentiment analysis, text summarization, and plagiarism detection. To facilitate an extension of the research on the document classification task, the data and code of our study are openly available\textsuperscript{\ref{lexicalchain-github}}.%TRV

%future work
%recurrent aspect
An exciting aspect of our architecture (Figure~\ref{fig:lc-arch}) is that FLLC II and FXLC II use synset embeddings models to decide how to represent their lexical chains. Once a synset embeddings model is available, using the results from FLLC II and FXLC II as a corpus, we can feed it back to our algorithms and create a more refined output corpus (using the original input one). This process can be done recurrently over many iterations, the same way MSSA-1R and MSSA-2R were generated in~\citep{Ruas:19}. We believe after many passes, the representation of our chains will get more accurate, better defined, and stable.

%F2F - Flex2Fixed
Considering FLLC II and FXLC II outputs have an identical format for their token representation (synset-annotated corpus), we can use them recurrently. Running the FLLC II algorithm iteratively will not affect its chain structure because once the flexible chains are represented, there is no lexical relationship that connects two separate chains. Otherwise, they would be placed together in the first place. However, for the FXLC II algorithm, if we use the output chain synset-annotated corpus as an input, the size of our corpus would decrease according to chunk size ($cs$). By doing so, FXLC II semantic representation would provide less meaningful chains.

%some ideas about our chains
In our techniques, we chose to represent each chain using the closest synset to their centroid, so we could still relate it to essential components in the lexical database. Therefore, less dominant synset are often not chosen as chain representatives. A clear direction would be to represent our chain as a direct vector-average of its constituents synsets (at the cost of some interpretability). Another alternative would be to use the average vector of the chain and look for the most similar(s) synsets in a pre-trained model. We leave the investigation of this and other variations of our algorithms for future work.

%final block - review
In this work, we chose to embed our lexical chains using a word2vec implementation. However, as explained in Section~\ref{sec:expe}, recently published embeddings techniques (e.g., ELMo, USE) bring new directions for our work. We believe our proposed approach of combining lexical chains and embeddings algorithms can leverage the semantic features in these neural network models. Moreover, we also intend to investigate how Transformer-based architectures \cite{Vaswani:17} such as BERT \cite{Devlin:18} and XLNet \cite{Yang:19} can be applied to FLLC and FXLC. However, these architectures will require additional effort in their implementation since their fine-tuning, training process, and hardware are more restrictive than traditional word embeddings techniques (e.g., word2vec, GloVe).

%% file: acknowledgment.tex
This work was partially supported by the Science Without Borders Brazilian Government Scholarship Program, CNPq  [grant number 205581/2014-5]; Charles Henrique Porto Ferreira was supported by Coordena\c{c}\~ao de Aperfei\c{c}oamento de Pessoal de N\'ivel Superior - Brasil (Capes), Programa de Doutorado Sandu\'iche no Exterior (PDSE), [grant number 88881.186965/2018-01].